\documentclass[pdflatex,sn-mathphys-num]{sn-jnl}% Math and Physical Sciences Numbered Reference Style 
\usepackage{dirtytalk}

\usepackage{graphicx}%
\usepackage{multirow}%
\usepackage{amsmath,amssymb,amsfonts}%
\usepackage{amsthm}%
\usepackage{mathrsfs}%
\usepackage[title]{appendix}%
\usepackage{textcomp}%
\usepackage{manyfoot}%
\usepackage{booktabs}%
\usepackage{algorithm}%
\usepackage{algorithmicx}%
\usepackage{algpseudocode}%
\usepackage{listings}%
\usepackage{subcaption}
\theoremstyle{thmstyleone}%
\theoremstyle{thmstyletwo}%

\theoremstyle{thmstylethree}%

\usepackage[table]{xcolor}% http://ctan.org/pkg/xcolor
\usepackage{placeins}
\usepackage[normalem]{ulem}
\useunder{\uline}{\ul}{}
\begin{document}
\title[Multi-Armed Bandit Approach for Optimizing Training on Synthetic Data]{Multi-Armed Bandit Approach for Optimizing Training on Synthetic Data}


\author*[1,2]{\fnm{Abdulrahman} \sur{Kerim}}\email{a.kerim@surrey.ac.uk}

\author[1]{\fnm{Leandro} \sur{Soriano Marcolino}}\email{l.marcolino@lancaster.ac.uk}

\author[3]{\fnm{Erickson} \sur{R. Nascimento}}\email{erickson@dcc.ufmg.br}

\author[1]{\fnm{Richard} \sur{Jiang}}\email{r.jiang2@lancaster.ac.uk}

\affil*[1]{\orgname{Lancaster University}, \orgaddress{\country{UK}}}

\affil[2]{\orgname{University of Surrey}, \orgaddress{\country{UK}}}

\affil[3]{\orgname{UFMG}, \orgaddress{\country{Brazil}}}

% \affil[4]{ \orgname{Microsoft}, \orgaddress{\country{Brazil}}}

% \affil[3]{\orgdiv{Department}, \orgname{Organization}, \orgaddress{\street{Street}, \city{City}, \postcode{610101}, \state{State}, \country{Country}}}


\abstract{Supervised machine learning methods require large-scale training datasets to perform well in practice. Synthetic data has been showing great progress recently and has been used as a complement to real data. However, there is yet a great urge to assess the usability of synthetically generated data. To this end, we propose a novel \textit{UCB}-based training procedure combined with a dynamic usability metric. Our proposed metric integrates low-level and high-level information from synthetic images and their corresponding real and synthetic datasets, surpassing existing traditional metrics. By utilizing a \textit{UCB}-based dynamic approach ensures continual enhancement of model learning. Unlike other approaches, our method effectively adapts to changes in the machine learning model's state and considers the evolving utility of training samples during the training process. We show that our metric is an effective way to rank synthetic images based on their usability. Furthermore, we propose a new attribute-aware bandit pipeline for generating synthetic data by integrating a Large Language Model with Stable Diffusion. Quantitative results show that our approach can boost the performance of a wide range of supervised classifiers. Notably, we observed an improvement of up to $10\%$ in classification accuracy compared to traditional approaches, demonstrating the effectiveness of our approach. Our source code, datasets, and additional materials are publically available at \url{https://github.com/A-Kerim/Synthetic-Data-Usability-2024}.}


\maketitle


\section{Introduction}
% [context] why we care about synthetic data in computer vision
The advancement in supervised \textit{Machine Learning (ML)} has been significantly influenced by the availability of large-scale annotated training data. State-of-the-art computer vision models~\cite{mettes2024hyperbolic, dubey2024transformer, xu2022groupvit, junayed2022himode, wang2023internimage} are typically trained on large-scale datasets. Traditional benchmarks such as \textit{MS COCO}~\cite{lin2014microsoft}, \textit{ADE20K}~\cite{zhou2019semantic, zhou2017scene}, \textit{CIFAR}~\cite{krizhevsky2009learning}, and \textit{ImageNet}~\cite{deng2009imagenet}, and more recent datasets like \textit{Panda-70m}~\cite{chen2024panda} and \textit{360+ x}~\cite{chen2024360+} have emerged to further push the boundaries of the field. However, collecting and annotating large-scale datasets is cumbersome and time-consuming, and the annotation process introduces privacy concerns, ethical issues, and the potential for biased and noisy human-labeled annotations. Synthetic data has emerged as an efficient solution for these problems~\cite{kerim2023synthetic,delussu2024synthetic}.  

% [problem] we can get a large-scale synthetic dataset but we do not know which samples are useful and which are not!
Synthetic data generation has seen remarkable advances in recent years, thanks to methods such as \textit{Generative Adversarial Networks}~\cite{iglesias2023survey, kang2023scaling,jin2020generative}, \textit{Diffusion Models}~\cite{nguyen2024dataset, croitoru2023diffusion, carlini2023extracting, bansal2023universal}, \textit{NeRF}~\cite{maxey2024uav,mildenhall2021nerf}, and simulators~\cite{dosovitskiy2017carla, shah2018airsim,ma2020new}. Advanced generation methods enable the creation of synthetic data which can augment, complement, or even replace real data in various computer vision applications~\cite{kerim2024leveraging, paulin2023review, delussu2024synthetic}. However, generated synthetic images are not always useful for training \textit{ML} models. Although humans can straightforwardly assess the photorealism and diversity of a small set of images, evaluating the usability of a large-scale synthetic dataset poses a significant challenge and might not even be directly correlated to photorealism and diversity in the first place.

% [Problem in more details] this is a primary  challenge but little work has been done!
Visually appealing images are not necessarily effective for training \textit{ML} models~\cite{tobin2017domain, nikolenko2021synthetic}. Similarly, highly diverse synthetic images are not certainly useful images, too~\cite{gong2019diversity}. A model trained on visually appealing images may struggle when faced with images that have variations in lighting and camera parameters which are common in practical scenarios but may not be present in aesthetically pleasing images. In contrast, overly diverse synthetic images may introduce unrealistic variations that do not reflect the true distribution of real-world data. Thus, it can confuse the model, making it less effective in practice.~\citeauthor{katayama2022domain, man2022review}~\cite{katayama2022domain, man2022review}  argue that a useful synthetic image is the one that is \textit{appropriately} both photorealistic and diverse. Photorealism is essential to bridge the domain gap between the synthetic and real domains and data diversity is fundamental for improving the generalizability and robustness of \textit{ML} models in practice~\cite{gong2019diversity, yang2022image, kariyappa2019improving}.

Therefore, it is crucial to allow the network to weigh the benefits of photorealism and diversity and dynamically select the best examples during training. A key contribution of our work is introducing a temporally adaptive parameter, evolving dynamically across epochs in response to the model’s internal state and current training context. This strategy is a key distinction in our approach compared to other metrics, which we demonstrate to be highly effective in optimizing training on synthetic data. 

While the need for this optimization is a primary challenge, there has been little work in computer vision literature. Traditional metrics such as \textit{Fréchet Inception Distance (FID)}~\cite{heusel2017gans} and \textit{Inception Score (IS)}~\cite{salimans2016improved}, have been widely used to evaluate the quality of generated synthetic images but not to assess their usability for training a ML model. This flexibility is essential for ensuring that the model learns a more representative and robust feature space, especially when dealing with heterogeneous data sources like synthetic and real-world images. Photorealistic examples enhance the model’s ability to generalize to real-world scenarios by providing higher fidelity visual information, while diverse examples help the model capture a wide range of variations in appearance, texture, and context. By prioritizing these types of examples, the network can avoid overfitting to less relevant or redundant data, thereby improving its performance across unseen data. Integrating a dynamic selection mechanism, such as through adaptive sampling techniques, ensures that the network focuses on the most informative samples at each stage of training, which leads to more efficient learning and stronger generalization capabilities across domains similar to our proposed approach.

The \textit{Upper Confidence Bound (UCB)} is a strategy used in decision-making processes, particularly in \textit{Reinforcement Learning (RL)} and \textit{Multi-Armed Bandit (MAB)} problems, to balance exploration and exploitation. In such scenarios, the goal is to maximize rewards by selecting actions or examples that are either well-understood (i.e., exploitation) or uncertain but potentially rewarding (i.e., exploration). \textit{UCB} achieves this by calculating an upper confidence bound for each action based on both the average reward and an uncertainty term. 

In this work, we utilize \textit{UCB} to identify and select the most valuable synthetic examples. Our approach balances exploration and exploitation by dynamically prioritizing synthetic samples that have higher uncertainty or higher potential impact on model learning. When training on both synthetic and real datasets, the utility of synthetic samples can vary significantly based on what is more essential at each training stage. To address this gap, we introduce a novel metric engineered to dynamically assess the utility of synthetic data during training, adapting to continuous changes in the synthetic data distribution and model state over time. Our proposed \textit{UCB}-based approach ensures that the model is not overly reliant on one data source while still maximizing the informative value of each training sample. By leveraging \textit{UCB}, the training process can adaptively choose the most useful examples from both datasets. This is particularly essential when the quality of synthetic data fluctuates or when there is a need to prioritize real data for critical learning stages, leading to more efficient and effective training. Hence, our main contributions are three-fold:

    \begin{itemize}
    \item[] \textbullet\hspace{\labelsep} We propose a dynamic adaptive metric for evaluating the usability of synthetic images;
    \item[] \textbullet\hspace{\labelsep}  We present a novel \textit{UCB}-based approach to optimize the training of supervised ML classifiers on synthetic data; 
    \item[] \textbullet\hspace{\labelsep} A new pipeline for automatically generating synthetic data by integrating a Large Language Model (LLM) with Stable Diffusion (SD);
    \item[] \textbullet\hspace{\labelsep}  Finally, we generate three synthetic artistic (i.e., \textit{SA-Car-2}, \textit{SA-CIFAR-10}, and \textit{SA-Birds-525}), and other three photorealistic synthetic datasets (i.e., \textit{SP-Car-2}, \textit{SP-CIFAR-10}, and \textit{SP-Birds-525}) for three classification tasks and we make them publicly available.
\end{itemize}

The rest of this paper is organized as follows. Section~\ref{Sec:Related Work} presents related work on synthetic data generation methods and usability metrics. Sub-Section~\ref{Sec:Problem Statement} articulates the problem statement and systematically formulates it. Sub-Section \ref{Sec:Attribute-Aware Synthetic Data Generation via Attribute-Aware Pipeline} introduces our synthetic data generation pipeline. The rest of Section~\ref{Sec:Methodology} explains the proposed usability metric, and the \textit{UCB}-based training procedure. In Section \ref{Sec:Experiments}, experimental results are demonstrated showing the effectiveness of our framework and approach. Section \ref{Sec:Conclusion} concludes the paper with a discussion of our work's limitations and potential future works.   
% The rest of this paper is organized as follows. Section 2 presents related work on xxxxx. Section 3 introduces uncertainty modeling and the UMSOT in detail. In Sect. 4, the motivation underlying our work and the proposed method and framework are evaluated with detailed experiments. Section 5 concludes the paper, and Sec. 6 presents the ethical statement.
%https://www.sciencedirect.com/science/article/abs/pii/S003442570400207X

\section{Related Work}
\label{Sec:Related Work}

In this section, we summarize key data generation methods and usability metrics. 
% and sample selection strategies. 
\subsection{Synthetic Data Generation Methods}
% Synthetic data generation methods have been showing great progress in the computer vision field. It seems as a promising solution to overcome the lack of suitable labeled data for training deep-supervised learning models~\cite{butler2012naturalistic,richter2016playing, ShafaeiLS16,liu2021urbanscene3d,kerim2021using,tsirikoglou2022synthetic}.
Several synthetic data generation methods have been developed recently. We review here the key approaches:

\noindent \textbf{Game Engines.}
The use of video games~\cite{ShafaeiLS16,kiefer2022leveraging} and game engines~\cite{kerim2021nova,lee2023game,wang2021pixel} has shown to be an effective way to generate automatically-labeled large-scale synthetic data for a wide range of computer vision tasks, such as semantic segmentation and depth estimation~\cite{ShafaeiLS16}, video stabilization~\cite{rao2023sim2realvs}, and person re-identification~\cite{sun2019dissecting}.
However, these approaches are limited in diversity and photorealism to game engines used, game genre, environment, and artists' skills. Additionally, creating large-scale datasets can be computationally expensive and poses intellectual property issues since video games are not usually developed for this aim~\cite{kerim2023synthetic}. Procedural content generation techniques may allow for the automatic creation of diverse and complex environments, reducing the need for manual design and increasing dataset variability~\cite{cobbe2020leveraging}. However, a key challenge remains in the domain gap between synthetic and real-world data. Despite advances in photorealism, discrepancies in sensor noise, motion blur, and subtle texture details often necessitate domain adaptation techniques~\cite{li2024bridging} or the incorporation of real data for fine-tuning models trained on synthetic data. To address these issues, hybrid approaches that combine synthetic and real-world data are being explored to enhance model generalization and reduce reliance on extensive labeled datasets~\cite{hao2024synthetic}. Despite these advances, selecting the most usable data from large-scale synthetic datasets remains underexplored as of yet.

\noindent \textbf{Generative Adversarial Networks (GANs).} GANs have been widely utilized to generate synthetic training data for computer vision~\cite{ali2023leveraging,torfi2022differentially}.~\citeauthor{al2023usability}~\cite{al2023usability} showed that utilizing synthetic images generated by their segmentation-informed conditional GAN improves the performance and robustness of heart cavity segmentation from short-axis cardiac magnetic resonance images. Similarly, it was shown in~\cite{li2023lift3d} that utilizing synthetic images generated by their GAN and Neural Radiance Field (NeRF)-based framework improves the performance of 3D object detectors. 
One of the primary challenges with these methods is their stability during training, frequently leading to issues such as mode collapse~\cite{kodali2017convergence}. This instability is frequently exacerbated by issues such as vanishing or exploding gradients, improper tuning of hyperparameters, and the sensitivity of the discriminator to slight perturbations in the generator’s output. Addressing these problems requires careful design choices, gradient penalty, improved weight initialization, and the use of robust loss functions that can stabilize the training dynamics~\cite{gulrajani2017improved, kang2023scaling}.

\noindent \textbf{Diffusion Models (DMs).}
Diffusion models generate data through a progressive denoising process, which refines random noise into realistic samples over multiple steps~\cite{croitoru2023diffusion,yang2023diffusion}. This iterative approach allows for better modeling of complex data distributions and produces outputs with higher fidelity and stability, making diffusion models increasingly preferred for tasks that demand robust and reliable generative performance. \textit{Stable Diffusion}, as demonstrated in~\cite{shipard2023diversity}, was used to boost the diversity of the generated images used for training a zero-shot classifier. Our work is similar to that work in using \textit{DMs} to generate the training images. However, our approach tackles the diversity and photorealism of the generated images. Furthermore, we incorporate an \textit{LLM} to create the attributes for the \textit{DMs'} prompts and we deploy a novel \textit{UCB}-based approach for selecting the best synthetic images for training.
% \erickson{Why using LLM for this, e.g., attribute creation, task could be a game changer?}. 
Training on synthetic images generated by \textit{DMs} for the task of skin disease classification was shown in~\cite{akrout2023diffusion} to achieve similar accuracy to training on real data. It was also shown that complementing \textit{ImageNet} training dataset with synthetic images generated by \textit{DMs} does improve \textit{ImageNet} classification accuracy substantially. 
Similar to that work, we also explore the effectiveness of training on synthetic data. However, instead of limiting our study to a single architecture, we consider six architectures: \textit{AlexNet}, \textit{EfficientNet}, \textit{ViT}, \textit{SwinTransformer}, \textit{VGG}, and \textit{REGNet}. These models were chosen to represent a diverse range of architectures. \textit{AlexNet} and \textit{VGG} are classical convolutional networks. \textit{EfficientNet} emphasizes model scaling for optimal accuracy-efficiency trade-offs. \textit{ViT} and \textit{SwinTransformer} leverage transformer-based architectures. Lastly, \textit{REGNet} represents a flexible architecture that utilizes dynamic scaling.

% \erickson{briefly list these six architectures and justify the motivation of selecting them}. 
% stable diffusion~\cite{shipard2023diversity, akrout2023diffusion, azizi2023synthetic}.

% \erickson{After reading this paragraph, I still don't get why the combination of LLM + diffusion models is a contribution. It is unclear why diffusion model and LLM.}
% Thus, the integration of LLMs with DMs models offers a holistic solution, addressing both semantic and visual aspects of synthetic image generation. This makes it a robust choice compared to other methods, particularly when aiming for diverse, realistic, and semantically aligned synthetic datasets for training computer vision models.

\subsection{Synthetic Data Usability}
There are many traditional metrics in the literature deployed to assess the quality of generated synthetic images, such as \textit{FID}~\cite{heusel2017gans}, \textit{IS}~\cite{salimans2016improved}, \textit{Peak-Signal-to-Noise Ratio} (\textit{PSNR}), and the \textit{Structural Similarity Index Measure} (\textit{SSIM})~\cite{wang2004image}. However, there are very few studies to assess the usability of generated synthetic images. 

Recent studies on the usability of synthetic data such as \cite{al2023usability, santangelo2024synthcheck, breuer2024validating, vallevik2024can} have a few limitations: a) they focus on the quality of the generated synthetic data not the usability; b) they are limited to certain narrow applications or fields. For instance, \cite{al2023usability} emphasises on the generation of semantically consistent and anatomically plausible images, yet these attributes alone do not guarantee that the synthetic data will effectively enhance the training of \textit{DL} models. This overlooks the necessity of ensuring that synthetic data captures the complex and varied characteristics required for robust model generalization across different clinical settings. On the other hand, \cite{santangelo2024synthcheck} focuses solely on generating and evaluating synthetic tabular data, while \textit{Electronic Health Records (EHRs)} often include other data types, such as bioimages and biosignals. Hence, the metrics implemented are specifically designed for tabular data, limiting their applicability to other data formats. 

% \erickson{Assessment versus usability. This relation is very confusing here. Maybe would be better "Assessment Usability of Synthetic Data"? }

\noindent \textbf{Traditional Synthetic Images Assessment Scores.} \textit{Fréchet Inception Distance (FID)} is usually used to assess the quality of the generated images by \textit{GANs}~\cite{lucic2018gans,borji2019pros}. It calculates the \textit{Wasserstein} distance between multivariate Gaussians embedded into a feature space of a specific layer of the \textit{Inception-V3} model~\cite{szegedy2016rethinking} pretrained on \textit{ImageNet}. \textit{IS}, for its turn, calculates the \textit{Kullback-Leibler (KL)} divergence between conditional and marginal class distributions utilizing the \textit{Inception-V3} model as well. Works such as~\cite{barratt2018note,ravuri2019seeing} have shown that \textit{FID} and \textit{IS} entangle diversity and photorealism (being close to the real distribution). While being useful for evaluating the performance of several models, neither metric is specifically designed to evaluate diversity within individual classes, which can be crucial in supervised learning. Thus, they may not be ideal metrics to assess the usability of synthetic images. 
% Additionally, they are not always interpretable in terms of photorealism and diversity.  
% \noindent \textbf{SSIM and PSNR.}
\textit{SSIM} can be used to measure the perceptual similarity between two images based on the degradation of structural information. On the other hand, \textit{PSNR} is the ratio between the original (real) image peak of power and the estimated power of the noise from synthetically generated or reconstructed image. ~\citeauthor{pambrun2015limitations}~\cite{pambrun2015limitations} and \citeauthor{huynh2008scope}~\cite{huynh2008scope} have shown that both metrics focus on pixel-level differences and are sensitive to simple geometric transformations. Additionally, they do not consider perceptual quality and diversity. 
Our proposed metric addresses these limitations by considering pixel-level and high-level information. Furthermore, we assess diversity and synthetic data feature consistency with real data counterparts.  
% \erickson{You are discussing some properties of your new metrics but still we don't have a glue about it. Maybe you could add a brief description in the introduction.}

\noindent \textbf{Global Consistency and Complexity Scores.}
~\citeauthor{scholz2023metrics}~\cite{scholz2023metrics} proposed two metrics to measure the global consistency of medical synthetic images (biological plausibility). They show that their approach is more robust compared to \textit{FID} as it can explicitly measure global consistency on a per-image basis. Unlike that work, our approach is a general-purpose metric, not limited to certain fields. At the same time, we evaluate the diversity of the generated images. Similar to our work,~\citeauthor{mahon2022measuring}~\cite{mahon2022measuring} propose a metric for measuring image complexity using hierarchical clustering. However, the authors make many assumptions about the cluster probability distribution.
% \erickson{which assumptions? And why is it a bad thing to have many assumptions? Are they wrong?}.
Additionally, it requires many hyperparameter tuning. 

% Unlike these metrics, ours
% % \erickson{approach = pipeline? or metric?} 
% takes into consideration low-level and high-level information about the synthetic image and its real and synthetic classes and datasets.  

Our proposed metric improves upon existing ones by incorporating both low-level and high-level information from synthetic images and their corresponding real and synthetic classes and datasets. Unlike traditional metrics, our \textit{UCB}-based dynamic approach enables ML models to select the most suitable examples for their current state at each epoch, ensuring that the training process is continuously optimized.
This key dynamic adaptability of our approach ensures that the model evolves with the data, enhancing its performance incrementally. This is in contrast to static metrics, which fail to account for the ongoing changes in the model's learning state and the varying utility of training samples throughout the training process.

% \erickson{The way you make a parallel between your metric and the available one are nice. However, the lack of information about the new metric works makes the discussion weak.}

% \subsection{Sample Selection Strategies}
% Effective sample selection is pivotal in training robust machine learning models. Traditional approaches often rely on heuristics or static sampling methods. Recent works have explored adaptive sampling techniques, such as Active Learning and Reinforcement Learning-based methods , to iteratively select the most informative samples for training. Multi-Armed Bandit (MAB) frameworks have also been employed to optimize sample selection, balancing exploration and exploitation to enhance model performance . Our work builds upon these strategies by integrating a UCB-based approach to dynamically select synthetic training samples, thereby improving training efficiency and effectiveness.

\section{Methodology}
\label{Sec:Methodology}

\begin{table*}[th]
\centering
\caption{Examples of the generated prompts used to create artistic (i.e., \textit{SA-Car-2}) and photorealistic (i.e., \textit{SP-Car-2}) versions of the \textit{Car Accidents} dataset. Each prompt includes specific attributes such as weather condition, accident type (if any), and vehicle color and model. The artistic prompts emphasize exaggerated stylistic effects, while the photorealistic prompts focus on capturing detailed, realistic imagery for each scenario.}
\label{tab:prompts}
\resizebox{\textwidth}{!}{%
\begin{tabular}{lll}
\hline
\multicolumn{1}{c}{Dataset} &
  \multicolumn{1}{c}{Selected Attributes} &
  \multicolumn{1}{c}{Prompts} \\ \hline
SA-Car-2 &
  \begin{tabular}[c]{@{}l@{}}weather: Thunderstorm,\\ accident: Head-on collision,\\ color: Brown,\\ model: Toyota Camry\end{tabular} &
  \begin{tabular}[c]{@{}l@{}} \say{Generate a highly stylized and non-photorealistic image of  a single car accident. \\ The accident type is Head-on collision occurring in Thunderstorm weather condition. \\ The car involved in the accident is of Brown color and is Toyota Camry model. \\ Apply unique and exaggerated artistic effects, such as vibrant color splashes, abstract shapes, and bold brushstrokes.}\end{tabular} \\ \hline
SA-Car-2 &
  \begin{tabular}[c]{@{}l@{}}weather: Dust storm,\\ accident: Rear-end collision,\\ color: Green,\\ model: BMW 3 Series\end{tabular} &
  \begin{tabular}[c]{@{}l@{}}\say{Generate a highly stylized and non-photorealistic image of a single car accident. \\ The accident type is Rear-end collision occurring in Dust storm weather condition. \\ The car involved in the accident is of Green color and is BMW 3 Series model. \\ Apply unique and exaggerated artistic effects, such as vibrant color splashes, abstract shapes, and bold brushstrokes.}\end{tabular} \\ \hline
SA-Car-2 &
  \begin{tabular}[c]{@{}l@{}}weather: Drizzle,\\ accident: Single-vehicle accident,\\ color: Yellow,\\ model: Volkswagen Golf\end{tabular} &
  \begin{tabular}[c]{@{}l@{}} \say{Generate a highly stylized and non-photorealistic image of a single car accident. \\ The accident type is Single-vehicle accident occurring in Drizzle weather condition. \\ The car involved in the accident is of Yellow color and is Volkswagen Golf model. \\ Apply unique and exaggerated artistic effects, such as vibrant color splashes, abstract shapes, and bold brushstrokes.}\end{tabular} \\ \hline
SA-Car-2 &
  \begin{tabular}[c]{@{}l@{}}weather: Partly cloudy,\\ accident: no-accident,\\ color: Blue,\\ model: Mercedes-Benz C-Class\end{tabular} &
  \begin{tabular}[c]{@{}l@{}} \say{Generate a highly stylized and non-photorealistic image of  a single car. The weather condition is Partly cloudy. \\ The car is of Blue color and is of Mercedes-Benz C-Class model. \\ Apply unique and exaggerated artistic effects, such as vibrant color splashes, abstract shapes, and bold brushstrokes.}\end{tabular} \\ \hline
SA-Car-2 &
  \begin{tabular}[c]{@{}l@{}}weather: Drizzle,\\ accident: no-accident,\\ color: Red,\\ model: Honda Civic\end{tabular} &
  \begin{tabular}[c]{@{}l@{}} \say{Generate a highly stylized and non-photorealistic image of  a single car. The weather condition is Drizzle. The car is of\\  Red color and is of Honda Civic model. Apply unique and exaggerated artistic effects, such as vibrant color splashes, \\ abstract shapes, and bold brushstrokes.}\end{tabular} \\ \hline 
SA-Car-2 &
  \begin{tabular}[c]{@{}l@{}}weather: Fog,\\ accident: no-accident,\\ color: Yellow,\\ model: Toyota Corolla\end{tabular} &
  \begin{tabular}[c]{@{}l@{}} \say{Generate a highly stylized and non-photorealistic image of  a single car. The weather condition is Fog.The car is of \\ Yellow color and is of Toyota Corolla model. Apply unique and exaggerated artistic effects, such as vibrant color\\  splashes, abstract shapes, and bold brushstrokes.}\end{tabular} \\ \hline  \hline
SP-Car-2 &
  \begin{tabular}[c]{@{}l@{}}weather: Snow,\\ accident: Improper lane change\\  collision,\\ color: Brown,\\ model: BMW 3 Series\end{tabular} &
  \begin{tabular}[c]{@{}l@{}} \say{Generate a photorealistic image of a single car accident. The accident type is Improper lane change collision\\  occurring in Snow weather condition.  The car involved in the accident is of Brown color and is BMW 3 Series model.\\ Capture the scene with meticulous attention to detail, realism, and visual impact.}\end{tabular} \\ \hline
SP-Car-2 &
  \begin{tabular}[c]{@{}l@{}}weather: Fog,\\ accident: Rear-end collision,\\ color: Silver,\\ model: Nissan Altima\end{tabular} &
  \begin{tabular}[c]{@{}l@{}} \say{Generate a photorealistic image of a single car accident. The accident type is Rear-end collision occurring in Fog \\ weather condition.  The car involved in the accident is of Silver color and is Nissan Altima model. Capture the scene \\ with meticulous attention to detail, realism, and visual impact.}\end{tabular} \\ \hline
SP-Car-2 &
  \begin{tabular}[c]{@{}l@{}}weather: Drizzle,\\ accident: Mechanical failure,\\ color: Brown,\\ model: Ford Mustang\end{tabular} &
  \begin{tabular}[c]{@{}l@{}} \say{Generate a photorealistic image of a single car accident. The accident type is Mechanical failure accident occurring in \\ Drizzle weather condition. The car involved in the accident is of Brown color and is Ford Mustang model. Capture the\\  scene with meticulous attention to detail, realism, and visual impact.}\end{tabular} \\ \hline
SP-Car-2 &
  \begin{tabular}[c]{@{}l@{}}weather: Sandstorm,\\ accident: No Accident,\\ color: Yellow,\\ model: Tesla Model 3\end{tabular} &
  \begin{tabular}[c]{@{}l@{}} \say{Generate a photorealistic image of a single car. The weather condition is Sandstorm. The car is of Yellow color and is of\\  Tesla Model 3 model. Capture the scene with meticulous attention to detail, realism, and visual impact.}\end{tabular} \\ \hline
SP-Car-2 &
  \begin{tabular}[c]{@{}l@{}}weather: Overcast,\\ accident: No Accident,\\ color: White,\\ model: Tesla Model 3\end{tabular} &
  \begin{tabular}[c]{@{}l@{}}\say{Generate a photorealistic image of a single car. The weather condition is Overcast. The car is of White color and is of\\  Tesla Model 3 model. Capture the scene with meticulous attention to detail, realism, and visual impact.}\end{tabular} \\ \hline
SP-Car-2 &
  \begin{tabular}[c]{@{}l@{}}weather: Tornado,\\ accident: No Accident,\\ color: Yellow,\\ model: Nissan Rogue\end{tabular} &
  \begin{tabular}[c]{@{}l@{}} \say{Generate a photorealistic image of a single car. The weather condition is Tornado. The car is of Yellow color and is of\\  Nissan Rogue model. Capture the scene with meticulous attention to detail, realism, and visual impact.}\end{tabular} \\ \hline
\end{tabular}%
}
\end{table*}

\subsection{Problem Statement}
\label{Sec:Problem Statement}
Let \(\mathcal{X}_{\text{real}} = \{x_i^{\text{real}}\}_{i=1}^n\) denote the set of real images and \(\mathcal{X}_{\text{syn}} = \{x_j^{\text{syn}}\}_{j=1}^m\) denote the set of synthetic images, where \(x_i^{\text{real}}\) and \(x_j^{\text{syn}}\) represent individual images from the real and synthetic domains, respectively. The photorealism of a synthetic image, \(x_j^{\text{syn}}\), can be quantified by measuring the distance between its corresponding features and the mean across the real set, \(\mathcal{X}_{\text{real}}\),  features in the latent space:
% \begin{equation}
% \text{Photorealism}(x_j^{\text{syn}}) =  d(\phi(x_j^{\text{syn}}), \phi(\mathcal{X}_{\text{real}})),   
% \end{equation}

\begin{equation}
\text{Photorealism}(x_j^{\text{syn}}) = d\left( \phi(x_j^{\text{syn}}), \frac{1}{|\mathcal{X}_{\text{real}}|} \sum_{x_i^{\text{real}} \in \mathcal{X}_{\text{real}}} \phi(x_i^{\text{real}}) \right),
\end{equation}

\noindent where \(\phi: x_i \rightarrow \mathbb{R}^k\) is an embedding function that maps images into a feature space \(\mathbb{R}^k\) and \(d(\cdot, \cdot)\) is a suitable distance metric (e.g., \(\ell_2\)-norm, Mahalanobis distance, and Cosine Similarity). It is also essential that the distribution \(\mathbb{P}(\mathcal{X}_{\text{syn}})\) approximates \(\mathbb{P}(\mathcal{X}_{\text{real}})\), ensuring, for instance, that \(\text{KL}(\mathbb{P}(\mathcal{X}_{\text{real}}) \parallel \mathbb{P}(\mathcal{X}_{\text{syn}}))\) is minimized, where \(\text{KL}(\cdot \parallel \cdot)\) represents the Kullback-Leibler divergence~\cite{kullback1951information}. Additionally, the diversity of \(\mathcal{X}_{\text{syn}}\) can be also assessed, for instance, through the entropy:
\begin{equation}
H(\mathbb{P}(\mathcal{X}_{\text{syn}})) = -\sum_{j=1}^m \mathbb{P}(x_j^{\text{syn}}) \log \mathbb{P}(x_j^{\text{syn}}).
\end{equation} 
% ~\cite{shannon1948mathematical, tuzovic2021mixing}
Therefore, to maximise the usability of \(\mathcal{X}_{\text{syn}}\), we need to maximise this joint objective function:

\begin{equation}
\mathcal{J}(\mathcal{X}_{\text{syn}}) = \alpha(t) \cdot \text{Photorealism}(\mathcal{X}_{\text{syn}}) + (1 -\alpha(t)) \cdot H(\mathbb{P}(\mathcal{X}_{\text{syn}})),
\end{equation}

\noindent where \(\alpha\) is hyperparameters that balance photorealism and diversity. 

This paper addresses the key challenge of leveraging synthetic data for training models that generalize well to real-world tasks, a problem driven by the domain gap between synthetic and real data. While synthetic data offers scalability and we can generate large-scale synthetic datasets, this gap often results in suboptimal performance on real-world applications even when trained on such large-scale datasets. Therefore, we propose an improved approach for generating synthetic images, evaluating their usability, and jointly training on synthetic and real data.

Our metric comprises two components: \textit{Diversity and Photorealism Score (DPS)} and \textit{Feature Cohesion Score (FCS)}. The \textit{DPS} evaluates the visual quality and diversity of the images, while the \textit{FCS} assesses the coherence of latent synthetic features in relation to their real counterpart. We also introduce a novel \textit{UCB}-based approach to dynamically select training samples during each epoch of the training process of a typical supervised ML model. This method not only enhances the selection process but also optimizes the model’s performance by focusing on the most beneficial training examples for a given ML model at each epoch. 

% \erickson{The current problem statement focuses primarily on describing your metric. It would be more effective to clearly articulate the specific problem your paper addresses. Consider expanding the discussion to include the broader context of the issue and how your work contributes to solving it.}

\newcommand{\realds}{\mathcal{R}}

% Symbol for synthetic dataset
\newcommand{\synthds}{\mathcal{S}}

% Parameters:
% M : number of best samples taken in the dataset 
% f_vgg() : feature extractor based on vgg
% DPS or DPS Term of Our Score (\Psi) 
% FCS or FCS Term of Our Score (\Phi)
% Our Score U : using UCB and selecting between DPS and FCS
% ME : mean(SSIM, PSNR, IS, FID, DPS, FCS, \textit{mean}(DPS, FCS))
% MD : median(SSIM, PSNR, IS, FID, DPS, FCS, \textit{mean}(DPS, FCS))
% MX : max(SSIM, PSNR, IS, FID, DPS, FCS, \textit{mean}(DPS, FCS))
% MN : min(SSIM, PSNR, IS, FID, DPS, FCS, \textit{mean}(DPS, FCS))

% I_i : synthetic image
% F_i : feature of image I_i generated by Fvgg or F_i = f_{vgg}(I_i)

% D\c: all classes in the dataset expect class c
% K : number of samples taken from each class for real data. 
% C  : number of classes in the dataset (synthetic or real)
% c : current class (synthetic or real)
% \textit{SA-Car-2} : Photorealistic synthetic car dataset
% \textit{SA-CIFAR-10} : Photorealistic synthetic cifar dataset
% \textit{SA-Birds-525} : Photorealistic synthetic birds dataset

% \[\bar{F} = \sum_i \frac{F_i}{K}\] : take the average of the features F_i

% \[\bar{F}_{c} = the average of the features for class c

% For discrete probability distributions
% D_{\text{KL}}(P \parallel Q) = \sum_{i} P(i) \log \frac{P(i)}{Q(i)}

% tau
% \upsilon
% \gamma
% \delta
% \mu
% \sigma : standard deviation

\subsection{Attribute-Aware Synthetic Data Generation via Attribute-Aware Pipeline}
\label{Sec:Attribute-Aware Synthetic Data Generation via Attribute-Aware Pipeline}
We propose a novel synthetic data generation pipeline utilising a \textit{Large Language Model (LLM)} with \textit{Diffusion Model (DM)} to create diverse and high-quality datasets. 
% ~\footnote{The synthetic data generator used in this work is part of a separate paper, which is currently under review. Details will be made available upon acceptance of the corresponding paper.}
The pipeline consists of three phases: Attribute extraction, Prompt creation and Image generation. Algorithm~\ref{alg:Synthetic Data-Generation-Pipeline} shows the details of each phase.

\noindent \textbf{a) Attribute Extraction using LLM.} 
% LLMs are pre-trained on large-scale and diverse datasets, making them a good fit for feature extraction.
% \erickson{Uncler which kind of feature you are talking about. Are they embedding vectors? Generally, LLMs do not provide embeddings. OpenAI doesn't provide them.}
% Thus, our methodology initiates by leveraging the advanced capabilities of LLMs to perform attribute extraction. Specifically, 
We employ an LLM to extract the main attributes of the dataset being generated. For instance, when generating car accidents synthetic dataset, we employ an LLM to identify the most popular car colors and models. 
% \erickson{which kind of conditions that are being considered as attribute here}.
This provides a pool of attributes for the second phase of the generation pipeline. 

\noindent \textbf{b) Attribute Sampling and Prompt Creation.} To construct the final prompts for the \textit{DM} to generate the required images, we randomly sample attributes from the pre-existing pool of attributes acquired in the previous phase. These sampled attributes are employed as input parameters to our \textit{DM} prompt template, which is designed to guide the \textit{DM} in generating realistic and contextually relevant data. The random sampling process ensures diversity in the generated datasets, thereby aiming at making them more representative of real-world scenarios. Table~\ref{tab:prompts} shows examples of the generated prompts for artistic and photorealistic versions of \textit{Car Accidents} datasets.  

% "aiming at making them more representative of real-world scenarios
% \erickson{The paragraph is only describing the attribute sampling. Where is the Prompt creation description? The community is very interested in prompt engineering, so here is a good place to elaborate on your strategy.}

\noindent \textbf{c) Data Generation with \textit{DM}.} At this stage, we leverage \textit{Stable Diffusion-V2} model~\cite{Rombach_2022_CVPR} to generate the required datasets using our carefully engineered prompts and pool of attributes generated using the \textit{LLM}.
% The DM was trained on \textit{LAION-5B} dataset~\cite{schuhmann2022laion} which contains more than $5.85$ billion image-text pairs to understand the context provided by the prompts and produce synthetic data that aligns with the specified attributes and conditions. 
% This extensive training enables the DM to effectively comprehend the contextual details encoded in the prompts, facilitating the generation of synthetic data that is more useful and practical.

% By leveraging the combined power of LLMs and DMs, we can efficiently generate high-quality data. This innovative and streamlined pipeline opens up new possibilities for training and evaluating machine learning models.

\begin{algorithm*}[htb]
\caption{Our Attribute-Aware Bandit Data Generation Pipeline Using LLM and SD}
\label{alg:Synthetic Data-Generation-Pipeline}
\begin{algorithmic}[1]

\Function{ExtractAttributes}{DomainContext, LLMModel}
    \State Initialize LLMModel
    \State Attributes $\gets$ LLMModel.\texttt{GenerateAttributes}(DomainContext)
    \State AttributePool $\gets$ \texttt{CompileAttributes}(Attributes)
    \State \Return AttributePool
\EndFunction
\State
\Function{CreatePrompts}{AttributePool, NumSamples, PromptTemplate}
    \State SampledAttributesList $\gets$ \texttt{RandomlySample}(AttributePool, NumSamples)
    \State Prompts $\gets$ \texttt{EmptyList}()
    \For{each SampledAttributes \textbf{in} SampledAttributesList}
        \State Prompt $\gets$ \texttt{FormatPrompt}(PromptTemplate, SampledAttributes)
        \State \texttt{Add Prompt to Prompts}()
    \EndFor
    \State \Return Prompts
\EndFunction
\State
\Function{GenerateImages}{Prompts, SDModel}
    \State Initialize SDModel
    \State Images $\gets$ \texttt{EmptyList}()
    \For{each Prompt \textbf{in} Prompts}
        \State Image $\gets$ SDModel.\texttt{GenerateImage}(Prompt)
        \State \texttt{Add Image to Images}()
    \EndFor
    \State Dataset $\gets$ \texttt{CompileImages}(Images)
    \State \Return Dataset
\EndFunction

\State 
\State AttributePool $\gets$ \Call{ExtractAttributes}{DomainContext, LLMModel} 
\State Prompts $\gets$ \Call{CreatePrompts}{AttributePool, NumSamples, PromptTemplate} 
\State SyntheticDataset $\gets$ \Call{GenerateImages}{Prompts, SDModel}

\end{algorithmic}
\end{algorithm*}

\begin{figure*}[htb]
  \centering
  \includegraphics[width=\linewidth]{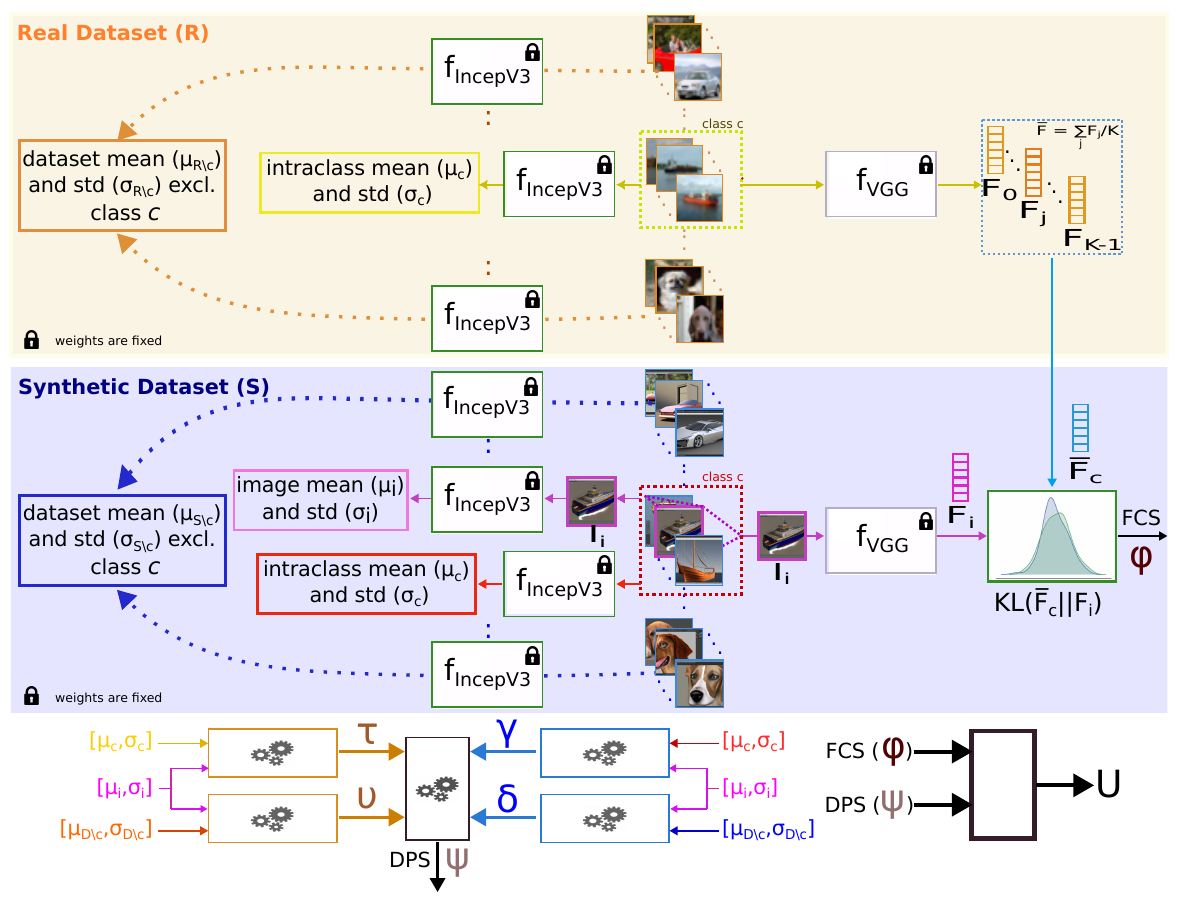}
   \caption{\textbf{Calculating Usability Score $\boldsymbol{U}$ for Synthetic Images.} Each synthetic image $I_i$ is assigned a usability score $\boldsymbol{U}$, which is derived from pixel-level and high-level information.}
   \label{fig:figure_2}
\end{figure*} 

\subsection{Usability Score}
Given the real dataset $\realds$ and its corresponding synthetic dataset $\synthds$, both comprising $C$ classes, we calculate the usability score as shown in Figure~\ref{fig:figure_2}. For each synthetic image $I_i$ from $\synthds$, we calculate its usability metric score $\mathbf{U}$ which is a $2$ dimensional vector: 
\begin{equation*}
 \mathbf{U} = \begin{bmatrix} \Psi& \Phi \end{bmatrix},
\end{equation*}

% -----------component 1
\noindent where $\Psi$, i.e., \textit{Diversity and Photorealism Score (DPS)}, represents the diversity and photorealism of $I_i$ considering only mid-level information. We utilise the \textit{Inception-V3} model~\cite{szegedy2016rethinking}, pre-trained on \textit{ImageNet}, to extract feature representations from the last convolution layer of the model for $I_i$:
\begin{equation*}
  V_i = f_{IncepV3}(I_i).
\end{equation*}
Then given $V_i$, the feature vector of image $I_i$, we calculate its mean $\mu_i$ and standard deviation $\sigma_i$. We also calculate the intraclass features mean $\mu_c$ and standard deviation $\sigma_c$ for the $I_i$ corresponding class $c$ from both $\realds$ and $\synthds$ datasets. Then, we calculate the synthetic $\synthds$ and real $\realds$ datasets features mean ($\mu_{S\backslash c}$ and $\mu_{R\backslash c}$) and standard deviation ($\sigma_{S\backslash c}$ and $\sigma_{R\backslash c}$) after excluding class $c$ images from $\realds$ and $\synthds$ datasets. 

% -----------component 2
% The second component, $\Phi$ or \textit{Feature Cohesion Score (FCS)}, represents the cohesion between high-level features of $I_i$ and the averaged real class features $\bar{F} = \sum_r \frac{F_r}{K}$, where for the synthetic image $I_i$ and each of the $K$ images from the class $c$ $F_r$ $\in$ $\realds$ dataset, $\realds_{k}^c$, we obtain a $4096$-dimensional feature vector:
The second component, $\Phi$, i.e., the \textit{Feature Cohesion Score (FCS)}, measures the cohesion between the high-level features of a synthetic image $I_i$ and the averaged features of the real images in the same class. The averaged real class features are computed as $\bar{F} = \frac{1}{K} \sum_r F_r$, where $F_r $ is a $4{,}096$-dimensional given by $F_r = f_{\text{vgg}}(I_r)$.

To compute $\bar{F}$ for a given real class $c$, $K$ real images from the class $c$ of the real dataset, $\realds_{k}^c$ are used. $K$ is a sampling parameter (i.e., a representation parameter) that controls how many real images are used to represent the real class. 
% \begin{equation*}
%   F_i = f_{vgg}(I_i)  
% \end{equation*}
% \begin{equation*} F_i = f_{\text{vgg}}(I_i) \end{equation*} \begin{equation*} F_j = f_{\text{vgg}}(R_j) \quad \text{for } j = 1, 2, \ldots, K \end{equation*}
% $F_i$ and $F_r$ are the feature vectors extracted by the \textit{VGG16} network for the synthetic image ($I_i$) and the real image ($I_r$), respectively.
% \begin{equation*} F_i = f_{\text{vgg}}(I_i) \end{equation*} 

% \begin{equation*} F_j^c = f_{\text{vgg}}(\realds_{k}^c) \quad \text{for } k = 1, 2, \ldots, K \end{equation*}
We use the \textit{VGG16} model that has been trained on the \textit{ImageNet} dataset as a feature extractor. 
Then, we calculate \textit{Kullback–Leibler} divergence between the normalized vector features, $\hat{F_i}$, of each synthetic image, $I_i$, and the normalized vector of the averaged features, $\hat{\bar{F}}$, of its corresponding class $c$ from the real dataset, $\realds$. Therefore, 
% \begin{equation}
%     \Phi = \frac{1}{D_{\text{KL}}(\hat{\bar{F}} \parallel \hat{F_i} )} = \frac{1}{\sum_{m=1}^{m=4096} \hat{\bar{F}}(m) *\log \frac{\hat{\bar{F}}(m)}{\hat{F_i}(m)}}
% \end{equation}
\begin{equation}
    \Phi = \frac{1}{D_{\text{KL}}(\hat{\bar{F}} \parallel \hat{F_i} )}.
\end{equation}
    % p (array-like): The first probability distribution (true distribution).
    % q (array-like): The second probability distribution (approximation).

\noindent \textbf{Intuition behind using \textit{VGG16} and \textit{Inception-V3}}: For \textit{FCS}, we utilize \textit{VGG16} with \textit{KL divergence} to compare the high-level features of synthetic images against the mean features of the corresponding real class. \textit{VGG16} was chosen because it can capture fine-grained, localized features, which makes it particularly suitable for highlighting the small distinctions between synthetic and real images within a class. Moreover, \textit{VGG16}'s sequential convolutional layers are very effective for extracting high-level representations that specifically focus on texture and detailed patterns, which are critical for assessing feature cohesion between synthetic and real data. For measuring diversity and photorealism (\textit{DPS}), we employ \textit{Inception-V3} for feature extraction. \textit{Inception-V3}'s architecture, with its multi-scale filters and broader receptive fields, is particularly effective at capturing both global and local patterns in images. This multi-scale capability allows it to better assess variations in image content, which is essential for evaluating diversity across a dataset. Moreover, \textit{Inception-V3} is designed to handle complex visual tasks and is known for its superior performance in identifying realistic image structures, making it more suitable for quantifying the photorealism of synthetic images.

\subsection{UCB-Based Training Procedure}
After calculating the usability scores and selecting the top $M$ images in $S$, we start fine-tuning the classification model on these images. Our approach leverages an improved version of \textit{UCB} algorithm to dynamically select the best data subset during the training process. The approach is outlined in Algorithm~\ref{alg:UCB-Algorithm}. First, we define a \textit{patience} threshold that determines how many epochs to wait for an improvement in validation accuracy before considering a switch in data subsets i.e., switch between best selected images according to $\Phi$ or $\Psi$. An initial arm, data subset, is then randomly selected from the available set of arms (two arms for our approach). This arm is used in the first epoch of the training loop. During each epoch $e$ of the total number of epochs, the model is fine-tuned using the \textit{best\_images} from the selected  \textit{current\_arm}. After fine-tuning for one epoch, we compute the validation accuracy for the model. The rewards for the current arm are then updated based on this validation accuracy.

We continuously check for improvements in the validation accuracy. If there is no improvement, we increment \textit{num\_epochs\_without\_improvement}. If this count exceeds the \textit{patience} threshold, we select a new arm using the \textit{select\_loader} function, which is based on the \textit{UCB} values. We compute the \textit{UCB} values for each loader (data subset) as:
\[
\text{ucb\_values} = \frac{\text{loader\_rewards}}{\text{loader\_counts} + \epsilon} + \beta \cdot \sqrt{\frac{\log(\text{total\_counts})}{\text{loader\_counts} + \epsilon}},
\]
\noindent where $\epsilon = 1e-5$ and $\beta=2$. Then, using \textit{UCB} values, we return the index of the loader with the highest \textit{UCB} value. This adaptive approach allows the model to be fine-tuned using the most useful synthetic images at each epoch tailored for the model and dataset.

\begin{figure}[t]
    \centering
    \includegraphics[width=1.0\linewidth]{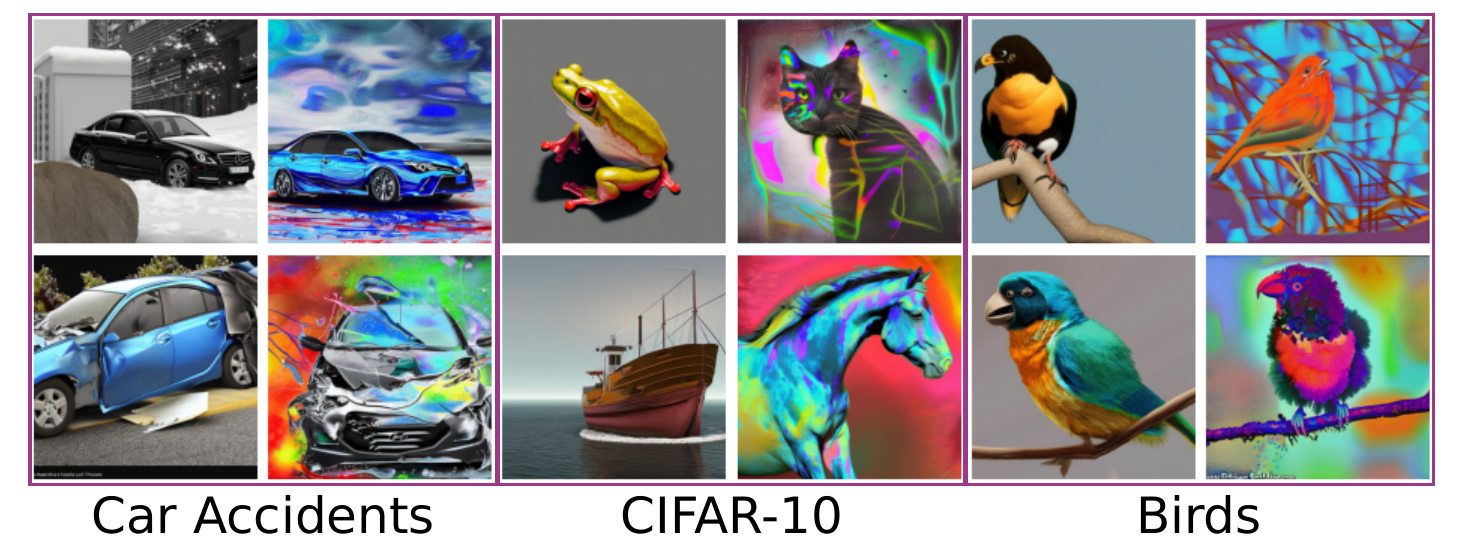}
    \caption{Samples from the synthetically generated datasets using our generation pipeline. Photorealistic and artistic samples are shown in the first and second columns of each dataset, respectively.}
    \label{fig:SynthDatasets}
\end{figure}

\begin{table*}[htb]
\caption{Synthetic statistics of the dataset generated by our approach.}
\label{tab:my-table}
\resizebox{\linewidth}{!}{%
\begin{tabular}{lllrr}
\hline
Task/Dataset & Artistic              & Photorealistic        & \# Classes & \# Images      \\ \hline
Car Accidents & \textit{SA-Car-2}     & \textit{SP-Car-2}     & $2$        & $844$      \\ \hline
CIFAR-10 &\textit{SA-CIFAR-10}  & \textit{SP-CIFAR-10}  & $10$       & $50,000$  \\ \hline
Birds & \textit{SA-Birds-525} & \textit{SP-Birds-525} & $525$      & $89,250$  \\ \hline
\end{tabular}
}
\end{table*}

\begin{algorithm*}[htb]
\caption{Our Proposed UCB-based Training Approach.}
\label{alg:UCB-Algorithm}
\begin{algorithmic}[1]

\Function{select\_loader}{}
    \If{$total\_counts < num\_loaders$}
        \State \Return $total\_counts$ \Comment{Initially select each loader once}
    \EndIf
    \State $ucb\_values \gets loader\_rewards / (loader\_counts + \epsilon) + \beta \cdot \sqrt{\log(total\_counts) / (loader\_counts + \epsilon)}$
    \State \Return $\arg\max(ucb\_values)$
\EndFunction
\State
\State $best\_val\_accuracy \gets 0$\Comment{ Initialization}
\State $num\_epochs\_without\_improvement \gets 0$
\State $patience \gets \text{patience\_threshold}$ 
% \Comment{Define the patience threshold}
\For{each image $I_i$ in $S$}\Comment{Get the scores of the synthetic images}
\State $I_i.\Phi \gets calculate\_phi(i)$
\State $I_i.\Psi \gets calculate\_psi(i)$
\EndFor

\State $best\_images \gets select\_top\_M\_images(S, M)$ \Comment{Filter the $M$ best examples}

\State $current\_arm \gets random\_select(arms)$\Comment{Initial arm selection for the first epoch}
\State
\For{each epoch $e$ in $total\_epochs$}\Comment{Training loop}
    \State $fine\_tune\_model(model, best\_images, current\_arm)$ \Comment{Fine-tune the model for one epoch}
 
    \State $val\_accuracy \gets calculate\_validation\_accuracy(model, validation\_set)$\Comment{Calculate the validation accuracy}
    
    \State $update\_rewards(current\_arm, val\_accuracy)$\Comment{Update UCB rewards}
    \State 
    \If{$val\_accuracy > best\_val\_accuracy$}\Comment{Check for improvement}
        \State $best\_val\_accuracy \gets val\_accuracy$
        \State $num\_epochs\_without\_improvement \gets 0$
    \Else
        \State $num\_epochs\_without\_improvement \ +=1$
    \EndIf
    
    \If{$num\_epochs\_without\_improvement > patience$}\Comment{Arm selection}
        \State $current\_arm \gets select\_loader()$
    \EndIf
\EndFor

\end{algorithmic}
\end{algorithm*}

\begin{table*}[htb]
\caption{Area Under the Curve (AUC) of accuracies for models trained from scratch using different proportions ($1\%$, $20\%$, $50\%$, $90\%$, and $100\%$) of non-photorealistic ($SA{\text-}[dataset \textunderscore name]$), photorealistic ($SP{\text-}[dataset \textunderscore name]$), and real datasets ($R{\text-}[dataset \textunderscore name]$). The \textbf{bold} values indicate the best performance achieved on each dataset, considering the three versions (artistic, photorealistic, and real), while the \underline{underlined} values represent the second-best performance.}
\label{tab:photorealism}
\resizebox{\textwidth}{!}{%
\begin{tabular}{l|ccc|ccc|ccc}
\hline
\textbf{}       & \multicolumn{3}{c|}{\textbf{Car Accidents}}          & \multicolumn{3}{c|}{\textbf{CIFAR-10}}               & \multicolumn{3}{c}{\textbf{Birds}}                   \\
                & \textit{SA-Car-2} & \textit{SP-Car-2} & \textit{R-Car-2}           & \textit{SA-CIFAR-10} & \textit{SP-CIFAR-10} &  \textit{R-CIFAR-10}           & \textit{SA-Birds-525} & \textit{SP-Birds-525} & \textit{R-Birds-525}           \\ \hline
\textbf{AlexNet }        & 55.58              & \textbf{76.10} & {\ul 68.71}    & 10.52              & {\ul 12.91}    & \textbf{58.44} & 0.99               & {\ul 1.99}     & \textbf{67.68} \\
\textbf{EfficientNet}    & 49.17              & {\ul 51.52}    & \textbf{58.73} & 9.57               & {\ul 9.64}     & \textbf{51.76} & 0.88               & {\ul 1.23}     & \textbf{36.88} \\
\textbf{ViT}             & 56.99              & {\ul 72.32}    & \textbf{79.22} & 12.08              & {\ul 15.68}    & \textbf{78.91} & 1.87               & {\ul 2.63}     & \textbf{58.59} \\
\textbf{SwinTransformer} & 49.50              & {\ul 64.06}    & \textbf{68.54} & 11.71              & {\ul 18.14}    & \textbf{42.40} & 1.78               & {\ul 1.91}     & \textbf{46.32} \\
\textbf{VGG}             & 53.44              & \textbf{85.32} & {\ul 82.93}    & 11.11              & {\ul 12.90}    & \textbf{88.75} & 0.39               & {\ul 0.44}     & \textbf{95.22} \\
\textbf{REGNet }         & 59.52              & \textbf{78.88} & {\ul 73.23}    & 10.07              & {\ul 10.42}    & \textbf{78.85} & 0.26               & {\ul 0.34}     & \textbf{78.03} \\\hline
\end{tabular}%
}
\end{table*}

\section{Experiments}
\label{Sec:Experiments}

\subsection{Experimental Setup}

\noindent \textbf{Synthetic Datasets.} We leverage our generation pipeline to generate six datasets for \textit{car accidents}, \textit{CIFAR-10}, and \textit{birds} classification problems. Samples from the datasets are shown in Figure~\ref{fig:SynthDatasets} and the statistics of the generated data are shown in Table~\ref{tab:my-table}. Image classification was selected as the downstream task because of the complexity of image classification tasks, which requires models to distinguish between a multitude of classes with varying degrees of intra-class variability and inter-class similarity, serves as a rigorous testbed for assessing the usability of our approach. Furthermore, image classification requires feature extraction and generalization capabilities. Thus, by training on a mixture of synthetic and real data, we can evaluate how effectively the model learns from the selected synthetic images using our approach versus others.

% \erickson{You should better motivate the chose of image classification as a downstream task to evaluate your approach. Is it challenging? Is there something in the problem that makes image classification a good representative task to show the benefits of using your method?}

% \noindent \textbf{Real Datasets.} We use three real datasets for training and benchmarking. 

% \noindent \textbf{Evaluation Metrics.}
% ......
% ......

% \noindent \textbf{Baselines.}
% ......
% ......

% \noindent \textbf{Implementation Details.} 
% ......
% ......

\subsection{Photorealism and Data Usability} 
We first analyze the importance of synthetic data photorealism on classification accuracy. We trained six architectures from scratch on three classification problems, each using non-photorealistic, photorealistic, and real datasets. We evaluated the performance by measuring the Area Under the Curve (AUC) of accuracies across varying dataset sizes ($1\%$, $20\%$, $50\%$, $90\%$, and $100\%$). Table~\ref{tab:photorealism} demonstrates that models trained on photorealistic data consistently achieved higher AUC values compared to those trained on non-photorealistic (artistic data), highlighting the domain gap problem where differences in data distribution between training and testing sets lead to poor performance on real data. Therefore, we can see that synthetic data realism is important for better usability.

\subsection{Evaluation of $\mathbf{U}$ Score with UCB-Based Training}
Our experiments demonstrate that the proposed $\boldsymbol{U}$ score, combined with our \textit{UCB}-based training approach, consistently achieves superior classification accuracy compared to other metrics. Table~\ref{tab:metric-comparison} provides a comparison of fine-tuning results for six architectures across the three photorealistic synthetic datasets: \textit{SP-Car-2}, \textit{SP-CIFAR-10}, and \textit{SP-Birds-525}. Architectures fine-tuned using our $\boldsymbol{U}$ score significantly outperformed those trained with alternative metrics. Our metric's ability to effectively select the most informative examples for training underlines its robustness and effectiveness. The \textit{UCB}-based training approach, by leveraging the $\boldsymbol{U}$ score, balances exploration and exploitation at each stage of the model's training, leading to improved model generalization and thus performance. The results clearly illustrate the advantage of our method in enhancing classification accuracy, underscoring its potential for broader application in fine-tuning neural network architectures.

\subsection{Qualitative Comparison Among Metrics}

\begin{figure}
\centering
\subcaptionbox{Top usable synthetic images from the \textit{SP-Car-2} dataset.}{%
  \includegraphics[width=1\textwidth]{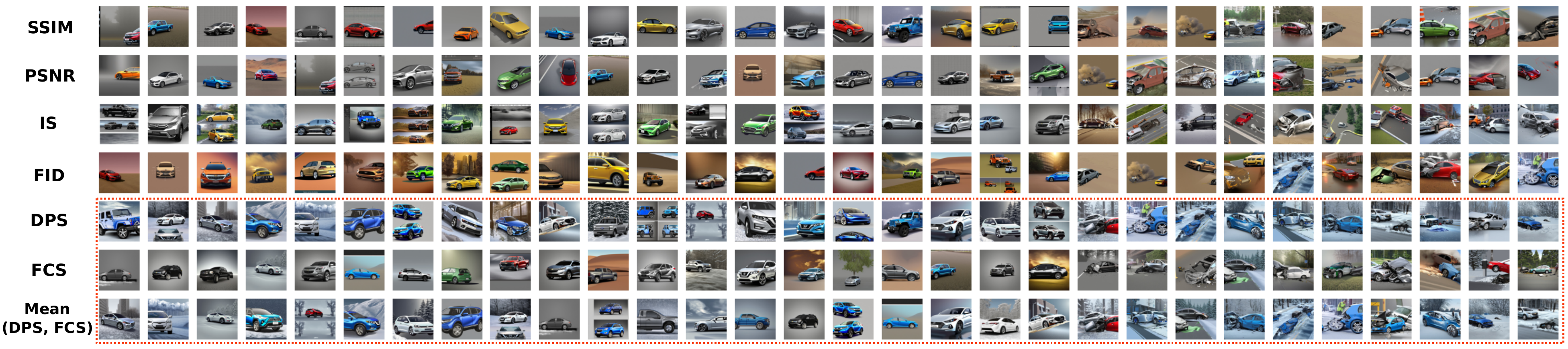}%
  }\par\medskip
\subcaptionbox{Top usable synthetic images from the \textit{SP-CIFAR-10} dataset.}{%
  \includegraphics[width=1\textwidth]{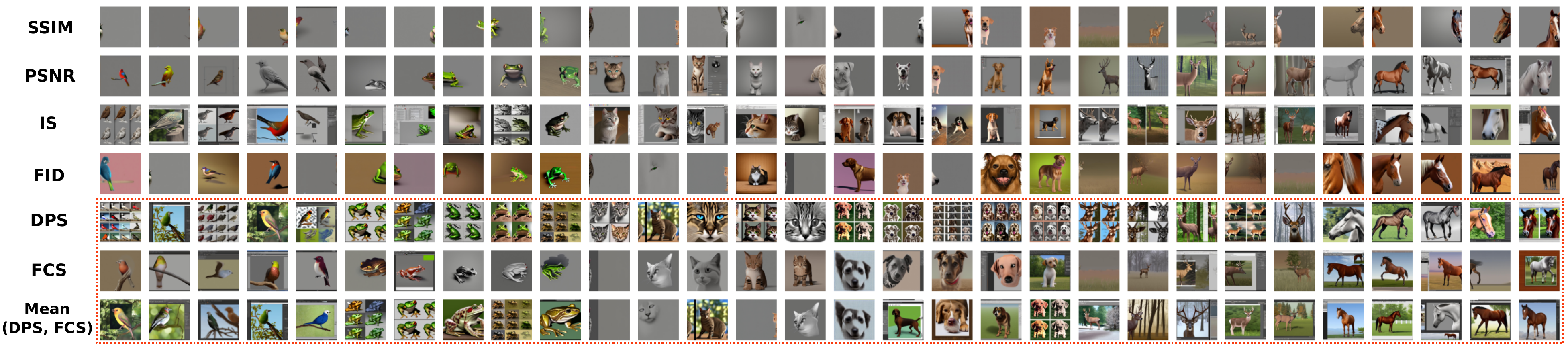}%
  }\par\medskip        
\subcaptionbox{Top usable synthetic images from the \textit{SP-Birds-525} dataset.}{%
  \includegraphics[width=1\textwidth]{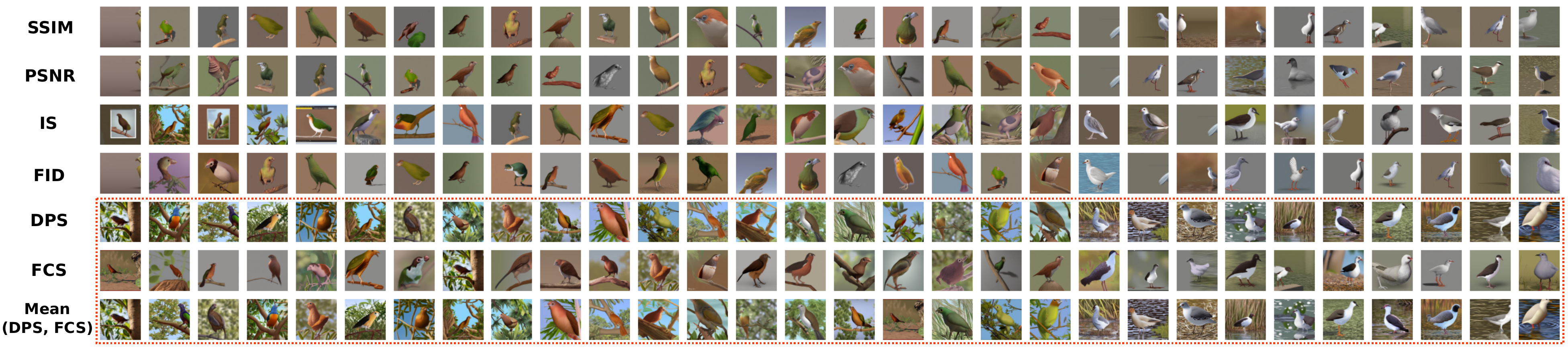}%
  }
\caption{\textbf{Top usable synthetic images on three synthetic datasets selected based on various metrics.} Traditional metrics struggle to consistently identify diverse and photorealistic images. In contrast, our approach (last row) effectively filters and highlights the most usable synthetic images. \textit{Best viewed in color and with zoom.}}
\label{fig:metric-comparison}
\end{figure}

To demonstrate that traditional metrics struggle to consistently identify diverse and photorealistic images, we show the top usable synthetic images according to these metrics and ours on three synthetic datasets in Figure~\ref{fig:metric-comparison}. Our proposed metric demonstrates an advantage in selecting more diverse and contextually relevant examples. For instance, images depicting snowy weather conditions were assigned higher priority, highlighting the ability of our approach to capture and emphasize rare but crucial environmental features. Additionally, unlike traditional metrics such as \textit{SSIM} and \textit{PSNR} which primarily focus on pixel-level similarity or overall image quality, our metric prioritizes more realistic backgrounds. This ensures that the selected samples are not only visually diverse but also semantically meaningful, leading to improved performance in downstream tasks as shown in Table~\ref{tab:metric-comparison}.

\subsection{Ablation Study} We conducted an ablation study to evaluate the contributions of different components in our framework shown in Table~\ref{tab:metric-ablation}. Initially, we assessed the classification accuracy by examining the \textit{DPS} ($\Psi$) and \textit{FCS} ($\Phi$) terms independently without the \textit{UCB}-based approach. This was followed by integrating these terms with \textit{UCB} to analyse their mutual effects. The results demonstrated that while \textit{DPS} and \textit{FCS} individually achieve good results, their combination with \textit{UCB} yielded significantly higher classification accuracy, highlighting the effectiveness of our proposed approach. We also report the results for using three arms (i.e., \textit{DPS}, \textit{FCS}, and the mean of \textit{DPS} and \textit{FCS}). The third arm in this setup serves as an approximation of our approach's behavior when the model is trained to convergence for a significant number of epochs. We also tried \textit{UCB} with traditional metrics such as \textit{SSIM}, \textit{PSNR}, \textit{IS}, and \textit{FID}. Finally, we utilized \textit{SSIM}, \textit{PSNR}, \textit{IS}, \textit{FID}, \textit{DPS}, and \textit{FCS} metrics. Additionally, we considered their aggregate statistics: mean (ME), median (MD), maximum (MX), and minimum (MN).   

\begin{figure}[ht]
    \begin{minipage}{0.515\linewidth}
        \subfloat[Percentage of common samples on the \textit{SP-Car-2} dataset.]{\label{fig:percentage-common-samples:a}\includegraphics[width=\linewidth]{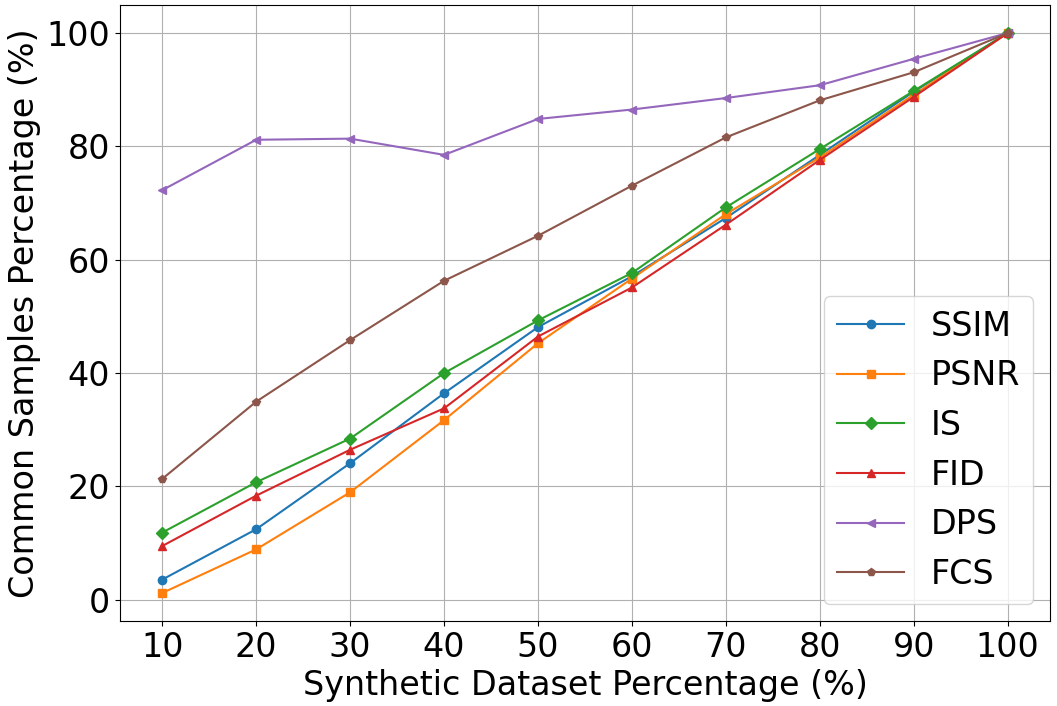}}
    \end{minipage}%
    \hfill
    \begin{minipage}{0.50\linewidth}
        \subfloat[Percentage of common samples on the \textit{SP-CIFAR-10} dataset.]{\label{fig:percentage-common-samples:b}\includegraphics[width=\linewidth]{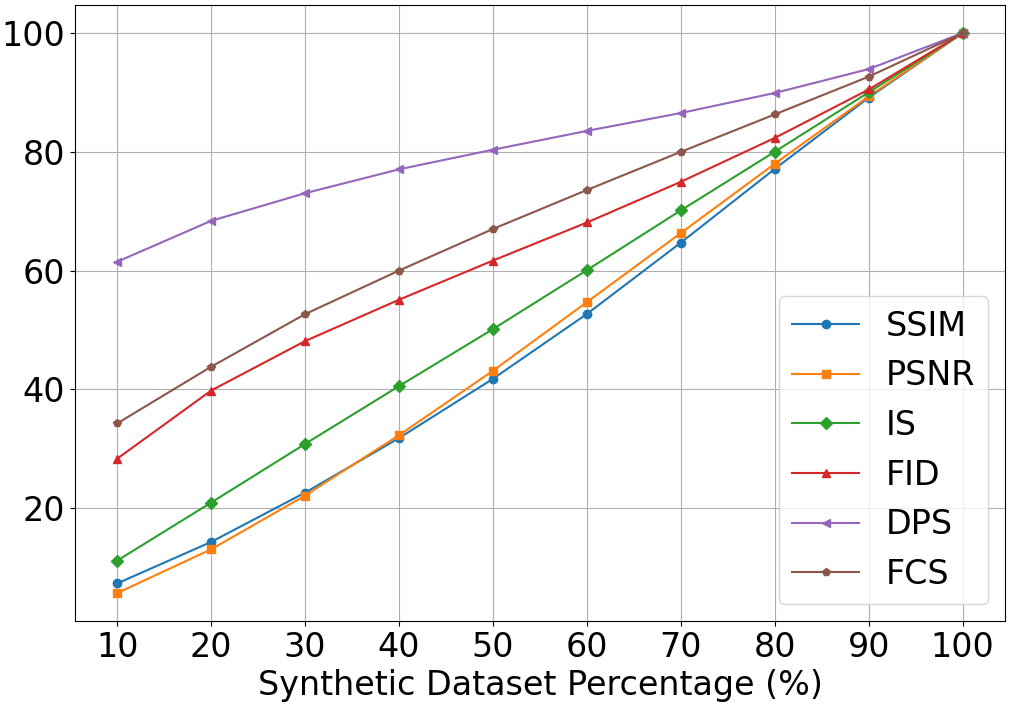}}
    \end{minipage}
    \par\medskip
    \centering
    \subfloat[Percentage of common samples on the \textit{SP-Birds-525} dataset.]{\label{fig:percentage-common-samples:c}\includegraphics[width=0.65\linewidth]{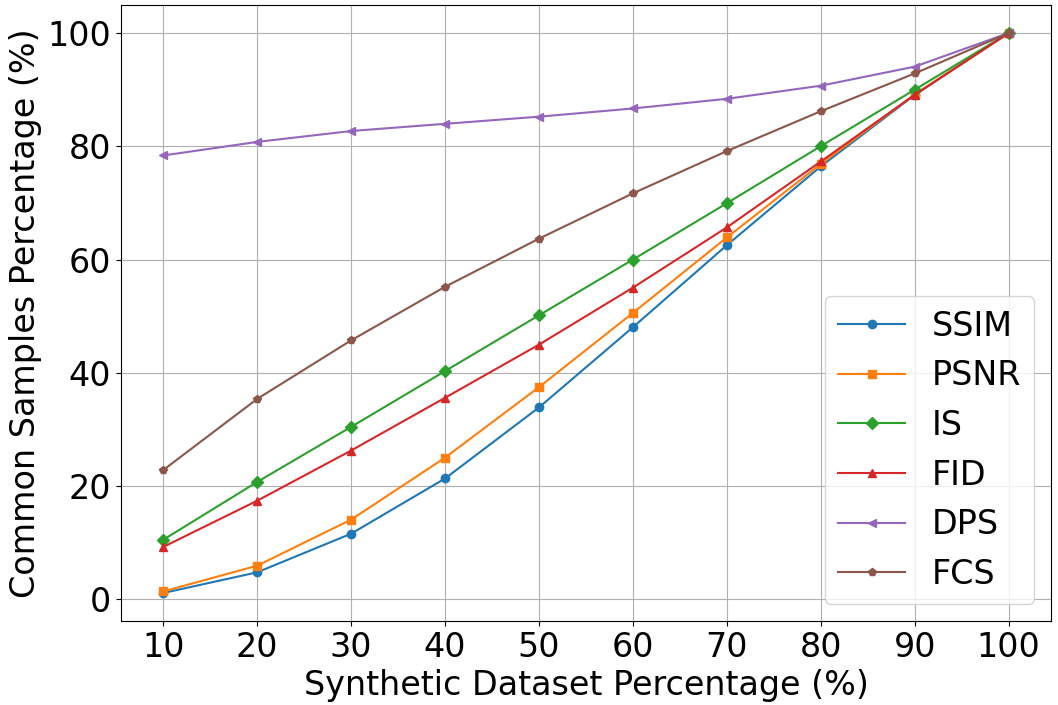}}

    \caption{Comparison of common examples identified by our metric (i.e., \textit{Mean(DPS, FCS)}) versus other metrics across three synthetic datasets: \textit{SP-Car-2}, \textit{SP-CIFAR-10}, and \textit{SP-Birds-525}.}
    \label{fig:percentage-common-samples}
\end{figure}

We also support our ablation study by rigorously examining the percentage of common samples identified by each metric. This analysis is conducted across varying dataset sizes on three synthetic datasets to ensure a comprehensive evaluation. The results clearly demonstrate that our metric consistently identifies distinct samples compared to traditional metrics, highlighting its capacity to capture different aspects of the data and provide more robust insights into model performance as shown in Figure~\ref{fig:percentage-common-samples}. 

\begin{table*}[]
\caption{\textbf{Classification Accuracy for Fine-Tuning Experiments Using our Approach.} This table shows the classification accuracy results of fine-tuning six architectures using half of the synthetic datasets: \textit{SP-Car-2}, \textit{SP-CIFAR-10}, and \textit{SP-Birds-525}. Various metrics, including ours, are employed to select the best examples. Our metric $\boldsymbol{U}$ with the proposed \textit{UCB-based} training approach achieves the best results.}
\label{tab:metric-comparison}
\resizebox{\textwidth}{!}{%
\begin{tabular}{ccccccc}
\toprule
                                   & \multicolumn{6}{c}{Car Accidents}                                                                                     \\ \hline
\multicolumn{1}{c|}{\textbf{}}     & \textbf{AlexNet} & \textbf{EfficientNet} & \textbf{ViT}  & \textbf{SwinTransformer} & \textbf{VGG}  & \textbf{REGNet} \\ \hline
\multicolumn{1}{c|}{SSIM}          & 73.0             & 64.0                  & 74.0          & {\ul 76.0}               & 66.0          & 77.0            \\
\multicolumn{1}{c|}{PSNR}          & {\ul 76.0}       & 63.0                  & {\ul 76.0}    & 73.0                     & 66.0          & 73.0            \\
\multicolumn{1}{c|}{IS}            & 74.0             & 64.0                  & 69.0          & {\ul 76.0}               & 67.0          & {\ul 78.0}      \\
\multicolumn{1}{c|}{FID}           & 71.0             & {\ul 65.0}            & 70.0          & 73.0                     & {\ul 71.0}    & 77.0            \\ 
\multicolumn{1}{c|}{Our Approach ($\boldsymbol{U})$} & \textbf{80.0}    & \textbf{75.0}         & \textbf{79.0} & \textbf{81.0}            & \textbf{77.0} & \textbf{85.0}   \\ \hline
                                   & \multicolumn{6}{c}{CIFAR-10}                                                                                             \\ \hline
\multicolumn{1}{c|}{\textbf{}}     & \textbf{AlexNet} & \textbf{EfficientNet} & \textbf{ViT}  & \textbf{SwinTransformer} & \textbf{VGG}  & \textbf{REGNet} \\ \hline
\multicolumn{1}{c|}{SSIM}          & 11.0             & 14.0                  & 21.0          & {\ul 30.0}               & 17.0          & {\ul 18.0}      \\
\multicolumn{1}{c|}{PSNR}          & {\ul 12.0}       & {\ul 16.0}            & {\ul 22.0}    & {\ul 30.0}               & {\ul 19.0}    & 17.0            \\
\multicolumn{1}{c|}{IS}            & 10.0             & 14.0                  & 21.0          & 28.0                     & 18.0          & 16.0            \\
\multicolumn{1}{c|}{FID}           & 11.0             & {\ul 16.0}            & 21.0          & 29.0                     & 18.0          & 15.0            \\
\multicolumn{1}{c|}{Our Approach ($\boldsymbol{U})$} & \textbf{19.0}    & \textbf{18.0}         & \textbf{28.0} & \textbf{33.0}            & \textbf{24.0} & \textbf{29.0}   \\ \hline
                                   & \multicolumn{6}{c}{Birds-525}                                                                                             \\ \hline
\multicolumn{1}{c|}{\textbf{}}                           & \textbf{AlexNet} & \textbf{EfficientNet} & \textbf{ViT}  & \textbf{SwinTransformer} & \textbf{VGG}  & \textbf{REGNet} \\ \hline
\multicolumn{1}{c|}{SSIM}          & 4.0              & {\ul $18.0$}            & {\ul 18.0}    & 14.0                     & {\ul 17.0}    & {\ul $16.0$}      \\
\multicolumn{1}{c|}{PSNR}          & {\ul $5.0$}        & {\ul $18.0$}            & {\ul 18.0}    & {\ul 15.0}               & 16.0          & {\ul $16.0$}      \\
\multicolumn{1}{c|}{IS}            & {\ul $5.0$}        & $17.0$                  & 17.0          & 14.0                     & 15.0          & {\ul $16.0$}      \\
\multicolumn{1}{c|}{FID}           & {\ul $5.0$}        & {\ul $18.0$}            & {\ul $18.0$}    & 14.0                     & {\ul 17.0}    & 15.0            \\
\multicolumn{1}{c|}{Our Approach ($\boldsymbol{U})$} & \textbf{10.0}    & \textbf{23.0}         & \textbf{28.0} & \textbf{23.0}            & \textbf{21.0} & \textbf{25.0}  \\ \bottomrule

\end{tabular}%
}
\end{table*}

\begin{table*}[]
\caption{\textbf{Ablation study results demonstrating the classification accuracy under different configurations.} We isolate the impact of the \textit{DPS} and \textit{FCS} terms individually without \textit{UCB}, then combine them with \textit{UCB}. We further evaluate the effect of incorporating three arms (\textit{DPS}, \textit{FCS}, and \textit{mean(DPS, FCS)}) and traditional metrics (\textit{SSIM}, \textit{PSNR}, \textit{IS}, \textit{FID}) with \textit{UCB}. Finally, we test the scalability with an 11-arm setup.  The shaded area indicates using our \textit{UCB-based} approach for training but with various number of arms.}
\label{tab:metric-ablation}
\resizebox{\textwidth}{!}{%
\begin{tabular}{cccccccc}
\hline
 & \multicolumn{6}{c}{Car Accidents} \\ \hline
\multicolumn{1}{c}{} & \multicolumn{1}{c|}{\textbf{}}   & \textbf{AlexNet} & \textbf{EfficientNet} & \textbf{ViT} & \textbf{SwinTransformer} & \textbf{VGG} & \textbf{REGNet} \\ \hline
\multicolumn{1}{c}{DPS Term of Our Score ($\Psi$)} & \multicolumn{1}{|c|}{1 Arm} & $67.0$ & $65.0$ & $69.0$ & $75.0$ & $62.0$ & $78.0$ \\
\multicolumn{1}{c}{FCS Term of Our Score ($\Phi$)}&  \multicolumn{1}{|c|}{1 Arm} & $71.0$ & $65.0$ & $74.0$ & $74.0$ & $67.0$ & $77.0$ \\
\multicolumn{1}{c}{\cellcolor{blue!6}Our Approach ($\boldsymbol{U})$}&  \multicolumn{1}{|c|}{2 Arms} & {\ul 80.0} & \textbf{75.0} & \textbf{79.0} & \textbf{81.0} & 77.0 & \textbf{85.0} \\
\multicolumn{1}{c}{\cellcolor{blue!6}DPS, FCS, \& \textit{mean}(DPS, FCS)} &  \multicolumn{1}{|c|}{3 Arms} & 73.0 & 71.0 & 72.0 & {\ul 79.0} & 77.0 & 79.0 \\
\multicolumn{1}{c}{\cellcolor{blue!6}SSIM, PSNR, IS, \& FID}&  \multicolumn{1}{|c|}{4 Arms} & \textbf{86.0} & 68.0 & 76.0 & 73.0 & \textbf{80.0} & 81.0 \\
\multicolumn{1}{c}{\begin{tabular}[c]{@{}c@{}}\cellcolor{blue!6}SSIM, PSNR, IS, FID, DPS, FCS, \textit{mean}(DPS, FCS), \\ \cellcolor{blue!6}ME, MD, MX, \& MN\end{tabular}} &  \multicolumn{1}{|c|}{11 Arms} & 79.0 & {\ul 73.0} & {\ul 78.0} & 69.0 & {\ul 79.0} & {\ul 83.0} \\ \hline

&\multicolumn{6}{c}{CIFAR-10} \\ \hline
\multicolumn{1}{c}{} & \multicolumn{1}{c|}{}  & \textbf{AlexNet} & \textbf{EfficientNet} & \textbf{ViT} & \textbf{SwinTransformer} & \textbf{VGG} & \textbf{REGNet} 
\\ \hline
\multicolumn{1}{c}{DPS Term of Our Score ($\Psi$)} &  \multicolumn{1}{|c|}{1 Arm} & $4.0$ & \textbf{$18.0$} & $17.0$ & $14.0$ & $16.0$ & $17.0$ \\
\multicolumn{1}{c}{FCS Term of Our Score ($\Phi$)} &  \multicolumn{1}{|c|}{1 Arm} & 11.0 & \textbf{18.0} & 23.0 & {\ul $30.0$} & 16.0 & 14.0 \\
\multicolumn{1}{c}{\cellcolor{blue!6}Our Approach ($\boldsymbol{U})$} &  \multicolumn{1}{|c|}{2 Arms} &\textbf{19.0} & \textbf{18.0} & \textbf{28.0} & \textbf{33.0} & \textbf{24.0} & \textbf{29.0} \\
\multicolumn{1}{c}{\cellcolor{blue!6}DPS, FCS, \& \textit{mean}(DPS, FCS)} & \multicolumn{1}{|c|}{3 Arms} & 11.0 & \textbf{18.0} & 22.0 & 28.0 & 16.0 & {\ul $20.0$} \\

\multicolumn{1}{c}{\cellcolor{blue!6}SSIM, PSNR, IS, \& FID}  &  \multicolumn{1}{|c|}{4 Arms} & {\ul 13.0} & {\ul 16.0} & {\ul 24.0} & 18.0 & 18.0 & 19.0 \\
\multicolumn{1}{c}{\begin{tabular}[c]{@{}c@{}}\cellcolor{blue!6}SSIM, PSNR, IS, FID, DPS, FCS, \textit{mean}(DPS, FCS), \\ \cellcolor{blue!6}ME, MD, MX, \& MN\end{tabular}}  &  \multicolumn{1}{|c|}{11 Arms}& 12.0 & {\ul 16.0} & 23.0 & 28.0 & {\ul 20.0} & 17.0 \\ \hline
 & \multicolumn{6}{c}{Birds-525} \\ \hline
\multicolumn{1}{c}{\textbf{}} &  \multicolumn{1}{c|}{\textbf{}} & \textbf{AlexNet} & \textbf{EfficientNet} & \textbf{ViT} & \textbf{SwinTransformer} & \textbf{VGG} & \textbf{REGNet} \\\hline
\multicolumn{1}{c}{DPS Term of Our Score ($\Psi$)} &  \multicolumn{1}{|c|}{1 Arm} & 4.0 & 18.0 & 17.0 & 14.0 & 16.0 & {\ul 17.0} \\
\multicolumn{1}{c}{FCS Term of Our Score ($\Phi$)} &  \multicolumn{1}{|c|}{1 Arm} & 4.0 & {\ul 20.0} & {\ul 19.0} & {\ul 16.0} & {\ul 18.0} & {\ul 17.0} \\
\multicolumn{1}{c}{\cellcolor{blue!6}Our Approach ($\boldsymbol{U})$} &  \multicolumn{1}{|c|}{2 Arms} &\textbf{10.0} & \textbf{23.0} & \textbf{28.0} & \textbf{23.0} & \textbf{21.0} & \textbf{25.0} \\
\multicolumn{1}{c}{\cellcolor{blue!6}DPS, FCS, \& \textit{mean}(DPS, FCS)} &  \multicolumn{1}{|c|}{3 Arms} & 4.0 & 15.0 & 16.0 & 11 & 17.0 & 13.0 \\
\multicolumn{1}{c}{\cellcolor{blue!6}SSIM, PSNR, IS, \& FID} &  \multicolumn{1}{|c|}{4 Arms} & {\ul 5.0} & 13.0 & 12.0 & 9.0 & 15.0 & 11.0 \\
\multicolumn{1}{c}{\begin{tabular}[c]{@{}c@{}}\cellcolor{blue!6}SSIM, PSNR, IS, FID, DPS, FCS, \textit{mean}(DPS, FCS), \\ \cellcolor{blue!6}ME, MD, MX, \& MN\end{tabular}} &  \multicolumn{1}{|c|}{11 Arms} & 4.0 & 16.0 & 15.0 & 10.0 & 16.0 & 15.0 \\ \hline
\end{tabular}%
}
\end{table*}

\FloatBarrier
\section{Conclusion}
\label{Sec:Conclusion}

In this work, we present a novel MAB-aware approach using \textit{UCB}-based training procedure and usability metric that offers a significant advancement in the evaluation and utilization of synthetic data for supervised machine learning and especially fundamental classification problems. By effectively integrating both low-level and high-level information, our metric provides a robust method for ranking synthetic images, leading to enhanced model performance. The introduction of a \textit{UCB}-based dynamic approach allows for continuous adaptation during training, ensuring optimal use of synthetic data. Our experimental results demonstrate that this approach not only complements real data but also significantly improves classification accuracy, highlighting its potential for widespread application in various supervised learning tasks.

\backmatter

% \bmhead{Supplementary information}

% \kr{If your article has accompanying supplementary file/s please state so here. }

% \kr{Authors reporting data from electrophoretic gels and blots should supply the full unprocessed scans for key as part of their Supplementary information. This may be requested by the editorial team/s if it is missing.}

% \kr{Please refer to Journal-level guidance for any specific requirements.}

% \bmhead{Acknowledgements}

% \kr{Acknowledgements are not compulsory. Where included they should be brief. Grant or contribution numbers may be acknowledged.
% Please refer to Journal-level guidance for any specific requirements.}

% \section*{Declarations}

% \kr{Some journals require declarations to be submitted in a standardised format. Please check the Instructions for Authors of the journal to which you are submitting to see if you need to complete this section. If yes, your manuscript must contain the following sections under the heading `Declarations':}

% \begin{itemize}
% \item Conflict of interest/Competing interests (check journal-specific guidelines for which heading to use)
% \item Ethics approval and consent to participate
% \item Consent for publication
% \item Materials availability
% \end{itemize}
\bigbreak
\noindent \textbf{Data Availability} Our source code, datasets, and additional materials are publically available at \url{https://github.com/A-Kerim/Synthetic-Data-Usability-2024}.

\bibliography{main}% common bib file


\begin{thebibliography}{71}
\ifx \bisbn   \undefined \def \bisbn  #1{ISBN #1}\fi
\ifx \binits  \undefined \def \binits#1{#1}\fi
\ifx \bauthor  \undefined \def \bauthor#1{#1}\fi
\ifx \batitle  \undefined \def \batitle#1{#1}\fi
\ifx \bjtitle  \undefined \def \bjtitle#1{#1}\fi
\ifx \bvolume  \undefined \def \bvolume#1{\textbf{#1}}\fi
\ifx \byear  \undefined \def \byear#1{#1}\fi
\ifx \bissue  \undefined \def \bissue#1{#1}\fi
\ifx \bfpage  \undefined \def \bfpage#1{#1}\fi
\ifx \blpage  \undefined \def \blpage #1{#1}\fi
\ifx \burl  \undefined \def \burl#1{\textsf{#1}}\fi
\ifx \doiurl  \undefined \def \doiurl#1{\url{https://doi.org/#1}}\fi
\ifx \betal  \undefined \def \betal{\textit{et al.}}\fi
\ifx \binstitute  \undefined \def \binstitute#1{#1}\fi
\ifx \binstitutionaled  \undefined \def \binstitutionaled#1{#1}\fi
\ifx \bctitle  \undefined \def \bctitle#1{#1}\fi
\ifx \beditor  \undefined \def \beditor#1{#1}\fi
\ifx \bpublisher  \undefined \def \bpublisher#1{#1}\fi
\ifx \bbtitle  \undefined \def \bbtitle#1{#1}\fi
\ifx \bedition  \undefined \def \bedition#1{#1}\fi
\ifx \bseriesno  \undefined \def \bseriesno#1{#1}\fi
\ifx \blocation  \undefined \def \blocation#1{#1}\fi
\ifx \bsertitle  \undefined \def \bsertitle#1{#1}\fi
\ifx \bsnm \undefined \def \bsnm#1{#1}\fi
\ifx \bsuffix \undefined \def \bsuffix#1{#1}\fi
\ifx \bparticle \undefined \def \bparticle#1{#1}\fi
\ifx \barticle \undefined \def \barticle#1{#1}\fi
\bibcommenthead
\ifx \bconfdate \undefined \def \bconfdate #1{#1}\fi
\ifx \botherref \undefined \def \botherref #1{#1}\fi
\ifx \url \undefined \def \url#1{\textsf{#1}}\fi
\ifx \bchapter \undefined \def \bchapter#1{#1}\fi
\ifx \bbook \undefined \def \bbook#1{#1}\fi
\ifx \bcomment \undefined \def \bcomment#1{#1}\fi
\ifx \oauthor \undefined \def \oauthor#1{#1}\fi
\ifx \citeauthoryear \undefined \def \citeauthoryear#1{#1}\fi
\ifx \endbibitem  \undefined \def \endbibitem {}\fi
\ifx \bconflocation  \undefined \def \bconflocation#1{#1}\fi
\ifx \arxivurl  \undefined \def \arxivurl#1{\textsf{#1}}\fi
\csname PreBibitemsHook\endcsname

%%% 1
\bibitem[\protect\citeauthoryear{Mettes et~al.}{2024}]{mettes2024hyperbolic}
\begin{botherref}
\oauthor{\bsnm{Mettes}, \binits{P.}},
\oauthor{\bsnm{Ghadimi~Atigh}, \binits{M.}},
\oauthor{\bsnm{Keller-Ressel}, \binits{M.}},
\oauthor{\bsnm{Gu}, \binits{J.}},
\oauthor{\bsnm{Yeung}, \binits{S.}}:
Hyperbolic deep learning in computer vision: A survey.
International Journal of Computer Vision,
1--25
(2024)
\end{botherref}
\endbibitem

%%% 2
\bibitem[\protect\citeauthoryear{Dubey and Singh}{2024}]{dubey2024transformer}
\begin{botherref}
\oauthor{\bsnm{Dubey}, \binits{S.R.}},
\oauthor{\bsnm{Singh}, \binits{S.K.}}:
Transformer-based generative adversarial networks in computer vision: A comprehensive survey.
IEEE Transactions on Artificial Intelligence
(2024)
\end{botherref}
\endbibitem

%%% 3
\bibitem[\protect\citeauthoryear{Xu et~al.}{2022}]{xu2022groupvit}
\begin{bchapter}
\bauthor{\bsnm{Xu}, \binits{J.}},
\bauthor{\bsnm{De~Mello}, \binits{S.}},
\bauthor{\bsnm{Liu}, \binits{S.}},
\bauthor{\bsnm{Byeon}, \binits{W.}},
\bauthor{\bsnm{Breuel}, \binits{T.}},
\bauthor{\bsnm{Kautz}, \binits{J.}},
\bauthor{\bsnm{Wang}, \binits{X.}}:
\bctitle{Groupvit: Semantic segmentation emerges from text supervision}.
In: \bbtitle{Proceedings of the IEEE/CVF Conference on Computer Vision and Pattern Recognition},
pp. \bfpage{18134}--\blpage{18144}
(\byear{2022})
\end{bchapter}
\endbibitem

%%% 4
\bibitem[\protect\citeauthoryear{Junayed et~al.}{2022}]{junayed2022himode}
\begin{bchapter}
\bauthor{\bsnm{Junayed}, \binits{M.S.}},
\bauthor{\bsnm{Sadeghzadeh}, \binits{A.}},
\bauthor{\bsnm{Islam}, \binits{M.B.}},
\bauthor{\bsnm{Wong}, \binits{L.-K.}},
\bauthor{\bsnm{Ayd{\i}n}, \binits{T.}}:
\bctitle{{HiMODE: A Hybrid Monocular Omnidirectional Depth Estimation Model}}.
In: \bbtitle{Proceedings of the IEEE/CVF Conference on Computer Vision and Pattern Recognition},
pp. \bfpage{5212}--\blpage{5221}
(\byear{2022})
\end{bchapter}
\endbibitem

%%% 5
\bibitem[\protect\citeauthoryear{Wang et~al.}{2023}]{wang2023internimage}
\begin{bchapter}
\bauthor{\bsnm{Wang}, \binits{W.}},
\bauthor{\bsnm{Dai}, \binits{J.}},
\bauthor{\bsnm{Chen}, \binits{Z.}},
\bauthor{\bsnm{Huang}, \binits{Z.}},
\bauthor{\bsnm{Li}, \binits{Z.}},
\bauthor{\bsnm{Zhu}, \binits{X.}},
\bauthor{\bsnm{Hu}, \binits{X.}},
\bauthor{\bsnm{Lu}, \binits{T.}},
\bauthor{\bsnm{Lu}, \binits{L.}},
\bauthor{\bsnm{Li}, \binits{H.}}, \betal:
\bctitle{{Internimage: Exploring large-scale vision foundation models with deformable convolutions}}.
In: \bbtitle{Proceedings of the IEEE/CVF Conference on Computer Vision and Pattern Recognition},
pp. \bfpage{14408}--\blpage{14419}
(\byear{2023})
\end{bchapter}
\endbibitem

%%% 6
\bibitem[\protect\citeauthoryear{Lin et~al.}{2014}]{lin2014microsoft}
\begin{bchapter}
\bauthor{\bsnm{Lin}, \binits{T.-Y.}},
\bauthor{\bsnm{Maire}, \binits{M.}},
\bauthor{\bsnm{Belongie}, \binits{S.}},
\bauthor{\bsnm{Hays}, \binits{J.}},
\bauthor{\bsnm{Perona}, \binits{P.}},
\bauthor{\bsnm{Ramanan}, \binits{D.}},
\bauthor{\bsnm{Doll{\'a}r}, \binits{P.}},
\bauthor{\bsnm{Zitnick}, \binits{C.L.}}:
\bctitle{{Microsoft COCO: Common Objects in Context}}.
In: \bbtitle{European Conference on Computer Vision}
(\byear{2014})
\end{bchapter}
\endbibitem

%%% 7
\bibitem[\protect\citeauthoryear{Zhou et~al.}{2019}]{zhou2019semantic}
\begin{barticle}
\bauthor{\bsnm{Zhou}, \binits{B.}},
\bauthor{\bsnm{Zhao}, \binits{H.}},
\bauthor{\bsnm{Puig}, \binits{X.}},
\bauthor{\bsnm{Xiao}, \binits{T.}},
\bauthor{\bsnm{Fidler}, \binits{S.}},
\bauthor{\bsnm{Barriuso}, \binits{A.}},
\bauthor{\bsnm{Torralba}, \binits{A.}}
\bjtitle{International Journal of Computer Vision}
\bvolume{127},
\bfpage{302}--\blpage{321}
(\byear{2019})
\end{barticle}
\endbibitem

%%% 8
\bibitem[\protect\citeauthoryear{Zhou et~al.}{2017}]{zhou2017scene}
\begin{bchapter}
\bauthor{\bsnm{Zhou}, \binits{B.}},
\bauthor{\bsnm{Zhao}, \binits{H.}},
\bauthor{\bsnm{Puig}, \binits{X.}},
\bauthor{\bsnm{Fidler}, \binits{S.}},
\bauthor{\bsnm{Barriuso}, \binits{A.}},
\bauthor{\bsnm{Torralba}, \binits{A.}}:
\bctitle{Scene parsing through ade20k dataset}.
In: \bbtitle{Proceedings of the IEEE Conference on Computer Vision and Pattern Recognition},
pp. \bfpage{633}--\blpage{641}
(\byear{2017})
\end{bchapter}
\endbibitem

%%% 9
\bibitem[\protect\citeauthoryear{Krizhevsky et~al.}{2009}]{krizhevsky2009learning}
\begin{botherref}
\oauthor{\bsnm{Krizhevsky}, \binits{A.}},
\oauthor{\bsnm{Hinton}, \binits{G.}}, et al.:
Learning multiple layers of features from tiny images
(2009)
\end{botherref}
\endbibitem

%%% 10
\bibitem[\protect\citeauthoryear{Deng et~al.}{2009}]{deng2009imagenet}
\begin{bchapter}
\bauthor{\bsnm{Deng}, \binits{J.}},
\bauthor{\bsnm{Dong}, \binits{W.}},
\bauthor{\bsnm{Socher}, \binits{R.}},
\bauthor{\bsnm{Li}, \binits{L.-J.}},
\bauthor{\bsnm{Li}, \binits{K.}},
\bauthor{\bsnm{Fei-Fei}, \binits{L.}}:
\bctitle{{ImageNet: A Large-Scale Hierarchical Image Database}}.
In: \bbtitle{2009 IEEE Conference on Computer Vision and Pattern Recognition}
(\byear{2009})
\end{bchapter}
\endbibitem

%%% 11
\bibitem[\protect\citeauthoryear{Chen et~al.}{2024a}]{chen2024panda}
\begin{bchapter}
\bauthor{\bsnm{Chen}, \binits{T.-S.}},
\bauthor{\bsnm{Siarohin}, \binits{A.}},
\bauthor{\bsnm{Menapace}, \binits{W.}},
\bauthor{\bsnm{Deyneka}, \binits{E.}},
\bauthor{\bsnm{Chao}, \binits{H.-w.}},
\bauthor{\bsnm{Jeon}, \binits{B.E.}},
\bauthor{\bsnm{Fang}, \binits{Y.}},
\bauthor{\bsnm{Lee}, \binits{H.-Y.}},
\bauthor{\bsnm{Ren}, \binits{J.}},
\bauthor{\bsnm{Yang}, \binits{M.-H.}}, \betal:
\bctitle{Panda-70m: Captioning 70m videos with multiple cross-modality teachers}.
In: \bbtitle{Proceedings of the IEEE/CVF Conference on Computer Vision and Pattern Recognition},
pp. \bfpage{13320}--\blpage{13331}
(\byear{2024})
\end{bchapter}
\endbibitem

%%% 12
\bibitem[\protect\citeauthoryear{Chen et~al.}{2024b}]{chen2024360+}
\begin{bchapter}
\bauthor{\bsnm{Chen}, \binits{H.}},
\bauthor{\bsnm{Hou}, \binits{Y.}},
\bauthor{\bsnm{Qu}, \binits{C.}},
\bauthor{\bsnm{Testini}, \binits{I.}},
\bauthor{\bsnm{Hong}, \binits{X.}},
\bauthor{\bsnm{Jiao}, \binits{J.}}:
\bctitle{360+ x: A panoptic multi-modal scene understanding dataset}.
In: \bbtitle{Proceedings of the IEEE/CVF Conference on Computer Vision and Pattern Recognition},
pp. \bfpage{19373}--\blpage{19382}
(\byear{2024})
\end{bchapter}
\endbibitem

%%% 13
\bibitem[\protect\citeauthoryear{Kerim}{2023}]{kerim2023synthetic}
\begin{bbook}
\bauthor{\bsnm{Kerim}, \binits{A.}}:
\bbtitle{Synthetic Data for Machine Learning: Revolutionize Your Approach to Machine Learning with this Comprehensive Conceptual Guide}.
\bpublisher{Packt Publishing},
\blocation{Birmingham, UK}
(\byear{2023}).
\burl{https://books.google.co.uk/books?id=JpXeEAAAQBAJ}
\end{bbook}
\endbibitem

%%% 14
\bibitem[\protect\citeauthoryear{Delussu et~al.}{2024}]{delussu2024synthetic}
\begin{botherref}
\oauthor{\bsnm{Delussu}, \binits{R.}},
\oauthor{\bsnm{Putzu}, \binits{L.}},
\oauthor{\bsnm{Fumera}, \binits{G.}}:
Synthetic data for video surveillance applications of computer vision: A review.
International Journal of Computer Vision,
1--37
(2024)
\end{botherref}
\endbibitem

%%% 15
\bibitem[\protect\citeauthoryear{Iglesias et~al.}{2023}]{iglesias2023survey}
\begin{barticle}
\bauthor{\bsnm{Iglesias}, \binits{G.}},
\bauthor{\bsnm{Talavera}, \binits{E.}},
\bauthor{\bsnm{D{\'\i}az-{\'A}lvarez}, \binits{A.}}:
\batitle{{A survey on GANs for computer vision: Recent research, analysis and taxonomy}}.
\bjtitle{Computer Science Review}
\bvolume{48},
\bfpage{100553}
(\byear{2023})
\end{barticle}
\endbibitem

%%% 16
\bibitem[\protect\citeauthoryear{Kang et~al.}{2023}]{kang2023scaling}
\begin{bchapter}
\bauthor{\bsnm{Kang}, \binits{M.}},
\bauthor{\bsnm{Zhu}, \binits{J.-Y.}},
\bauthor{\bsnm{Zhang}, \binits{R.}},
\bauthor{\bsnm{Park}, \binits{J.}},
\bauthor{\bsnm{Shechtman}, \binits{E.}},
\bauthor{\bsnm{Paris}, \binits{S.}},
\bauthor{\bsnm{Park}, \binits{T.}}:
\bctitle{{Scaling up GANs for text-to-image synthesis}}.
In: \bbtitle{Proceedings of the IEEE/CVF Conference on Computer Vision and Pattern Recognition},
pp. \bfpage{10124}--\blpage{10134}
(\byear{2023})
\end{bchapter}
\endbibitem

%%% 17
\bibitem[\protect\citeauthoryear{Jin et~al.}{2020}]{jin2020generative}
\begin{botherref}
\oauthor{\bsnm{Jin}, \binits{L.}},
\oauthor{\bsnm{Tan}, \binits{F.}},
\oauthor{\bsnm{Jiang}, \binits{S.}}, et al.:
Generative adversarial network technologies and applications in computer vision.
Computational intelligence and neuroscience
\textbf{2020}
(2020)
\end{botherref}
\endbibitem

%%% 18
\bibitem[\protect\citeauthoryear{Nguyen et~al.}{2024}]{nguyen2024dataset}
\begin{botherref}
\oauthor{\bsnm{Nguyen}, \binits{Q.}},
\oauthor{\bsnm{Vu}, \binits{T.}},
\oauthor{\bsnm{Tran}, \binits{A.}},
\oauthor{\bsnm{Nguyen}, \binits{K.}}:
Dataset diffusion: Diffusion-based synthetic data generation for pixel-level semantic segmentation.
Advances in Neural Information Processing Systems
\textbf{36}
(2024)
\end{botherref}
\endbibitem

%%% 19
\bibitem[\protect\citeauthoryear{Croitoru et~al.}{2023}]{croitoru2023diffusion}
\begin{barticle}
\bauthor{\bsnm{Croitoru}, \binits{F.-A.}},
\bauthor{\bsnm{Hondru}, \binits{V.}},
\bauthor{\bsnm{Ionescu}, \binits{R.T.}},
\bauthor{\bsnm{Shah}, \binits{M.}}:
\batitle{Diffusion models in vision: A survey}.
\bjtitle{IEEE Transactions on Pattern Analysis and Machine Intelligence}
\bvolume{45}(\bissue{9}),
\bfpage{10850}--\blpage{10869}
(\byear{2023})
\end{barticle}
\endbibitem

%%% 20
\bibitem[\protect\citeauthoryear{Carlini et~al.}{2023}]{carlini2023extracting}
\begin{bchapter}
\bauthor{\bsnm{Carlini}, \binits{N.}},
\bauthor{\bsnm{Hayes}, \binits{J.}},
\bauthor{\bsnm{Nasr}, \binits{M.}},
\bauthor{\bsnm{Jagielski}, \binits{M.}},
\bauthor{\bsnm{Sehwag}, \binits{V.}},
\bauthor{\bsnm{Tramer}, \binits{F.}},
\bauthor{\bsnm{Balle}, \binits{B.}},
\bauthor{\bsnm{Ippolito}, \binits{D.}},
\bauthor{\bsnm{Wallace}, \binits{E.}}:
\bctitle{Extracting training data from diffusion models}.
In: \bbtitle{32nd USENIX Security Symposium (USENIX Security 23)},
pp. \bfpage{5253}--\blpage{5270}
(\byear{2023})
\end{bchapter}
\endbibitem

%%% 21
\bibitem[\protect\citeauthoryear{Bansal et~al.}{2023}]{bansal2023universal}
\begin{bchapter}
\bauthor{\bsnm{Bansal}, \binits{A.}},
\bauthor{\bsnm{Chu}, \binits{H.-M.}},
\bauthor{\bsnm{Schwarzschild}, \binits{A.}},
\bauthor{\bsnm{Sengupta}, \binits{S.}},
\bauthor{\bsnm{Goldblum}, \binits{M.}},
\bauthor{\bsnm{Geiping}, \binits{J.}},
\bauthor{\bsnm{Goldstein}, \binits{T.}}:
\bctitle{Universal guidance for diffusion models}.
In: \bbtitle{Proceedings of the IEEE/CVF Conference on Computer Vision and Pattern Recognition},
pp. \bfpage{843}--\blpage{852}
(\byear{2023})
\end{bchapter}
\endbibitem

%%% 22
\bibitem[\protect\citeauthoryear{Maxey et~al.}{2024}]{maxey2024uav}
\begin{bchapter}
\bauthor{\bsnm{Maxey}, \binits{C.}},
\bauthor{\bsnm{Choi}, \binits{J.}},
\bauthor{\bsnm{Lee}, \binits{H.}},
\bauthor{\bsnm{Manocha}, \binits{D.}},
\bauthor{\bsnm{Kwon}, \binits{H.}}:
\bctitle{{UAV-Sim: NeRF-based Synthetic Data Generation for UAV-based Perception}}.
In: \bbtitle{2024 IEEE International Conference on Robotics and Automation (ICRA)},
pp. \bfpage{5323}--\blpage{5329}
(\byear{2024}).
\bcomment{IEEE}
\end{bchapter}
\endbibitem

%%% 23
\bibitem[\protect\citeauthoryear{}{}]{mildenhall2021nerf}
\begin{botherref}
{NeRF: Representing Scenes as Neural Radiance Fields for View Synthesis}
\end{botherref}
\endbibitem

%%% 24
\bibitem[\protect\citeauthoryear{Dosovitskiy et~al.}{2017}]{dosovitskiy2017carla}
\begin{bchapter}
\bauthor{\bsnm{Dosovitskiy}, \binits{A.}},
\bauthor{\bsnm{Ros}, \binits{G.}},
\bauthor{\bsnm{Codevilla}, \binits{F.}},
\bauthor{\bsnm{Lopez}, \binits{A.}},
\bauthor{\bsnm{Koltun}, \binits{V.}}:
\bctitle{{CARLA: An open urban driving simulator}}.
In: \bbtitle{Conference on Robot Learning},
pp. \bfpage{1}--\blpage{16}
(\byear{2017}).
\bcomment{PMLR}
\end{bchapter}
\endbibitem

%%% 25
\bibitem[\protect\citeauthoryear{Shah et~al.}{2018}]{shah2018airsim}
\begin{bchapter}
\bauthor{\bsnm{Shah}, \binits{S.}},
\bauthor{\bsnm{Dey}, \binits{D.}},
\bauthor{\bsnm{Lovett}, \binits{C.}},
\bauthor{\bsnm{Kapoor}, \binits{A.}}:
\bctitle{Airsim: High-fidelity visual and physical simulation for autonomous vehicles}.
In: \bbtitle{Field and Service Robotics},
pp. \bfpage{621}--\blpage{635}
(\byear{2018}).
\bcomment{Springer}
\end{bchapter}
\endbibitem

%%% 26
\bibitem[\protect\citeauthoryear{Ma et~al.}{2020}]{ma2020new}
\begin{bchapter}
\bauthor{\bsnm{Ma}, \binits{C.}},
\bauthor{\bsnm{Zhou}, \binits{Y.}},
\bauthor{\bsnm{Li}, \binits{Z.}}:
\bctitle{{A New Simulation Environment Based on Airsim, ROS, and PX4 for Quadcopter Aircrafts}}.
In: \bbtitle{2020 6th International Conference on Control, Automation and Robotics (ICCAR)},
pp. \bfpage{486}--\blpage{490}
(\byear{2020}).
\bcomment{IEEE}
\end{bchapter}
\endbibitem

%%% 27
\bibitem[\protect\citeauthoryear{Kerim et~al.}{2024}]{kerim2024leveraging}
\begin{bchapter}
\bauthor{\bsnm{Kerim}, \binits{A.}},
\bauthor{\bsnm{Ramos}, \binits{W.L.}},
\bauthor{\bsnm{Marcolino}, \binits{L.S.}},
\bauthor{\bsnm{Nascimento}, \binits{E.R.}},
\bauthor{\bsnm{Jiang}, \binits{R.}}:
\bctitle{Leveraging synthetic data to learn video stabilization under adverse conditions}.
In: \bbtitle{Proceedings of the IEEE/CVF Winter Conference on Applications of Computer Vision},
pp. \bfpage{6931}--\blpage{6940}
(\byear{2024})
\end{bchapter}
\endbibitem

%%% 28
\bibitem[\protect\citeauthoryear{Paulin and Ivasic-Kos}{2023}]{paulin2023review}
\begin{barticle}
\bauthor{\bsnm{Paulin}, \binits{G.}},
\bauthor{\bsnm{Ivasic-Kos}, \binits{M.}}:
\batitle{Review and analysis of synthetic dataset generation methods and techniques for application in computer vision}.
\bjtitle{Artificial intelligence review}
\bvolume{56}(\bissue{9}),
\bfpage{9221}--\blpage{9265}
(\byear{2023})
\end{barticle}
\endbibitem

%%% 29
\bibitem[\protect\citeauthoryear{Tobin et~al.}{2017}]{tobin2017domain}
\begin{bchapter}
\bauthor{\bsnm{Tobin}, \binits{J.}},
\bauthor{\bsnm{Fong}, \binits{R.}},
\bauthor{\bsnm{Ray}, \binits{A.}},
\bauthor{\bsnm{Schneider}, \binits{J.}},
\bauthor{\bsnm{Zaremba}, \binits{W.}},
\bauthor{\bsnm{Abbeel}, \binits{P.}}:
\bctitle{{Domain randomization for transferring deep neural networks from simulation to the real world}}.
In: \bbtitle{2017 IEEE/RSJ International Conference on Intelligent Robots and Systems (IROS)},
pp. \bfpage{23}--\blpage{30}
(\byear{2017}).
\bcomment{IEEE}
\end{bchapter}
\endbibitem

%%% 30
\bibitem[\protect\citeauthoryear{Nikolenko}{2021}]{nikolenko2021synthetic}
\begin{bbook}
\bauthor{\bsnm{Nikolenko}, \binits{S.I.}}:
\bbtitle{{Synthetic Data for Deep Learning}}
vol. \bseriesno{174}.
\bpublisher{Springer}, \blocation{???}
(\byear{2021})
\end{bbook}
\endbibitem

%%% 31
\bibitem[\protect\citeauthoryear{Gong et~al.}{2019}]{gong2019diversity}
\begin{barticle}
\bauthor{\bsnm{Gong}, \binits{Z.}},
\bauthor{\bsnm{Zhong}, \binits{P.}},
\bauthor{\bsnm{Hu}, \binits{W.}}:
\batitle{{Diversity in Machine Learning}}.
\bjtitle{IEEE Access}
\bvolume{7},
\bfpage{64323}--\blpage{64350}
(\byear{2019})
\end{barticle}
\endbibitem

%%% 32
\bibitem[\protect\citeauthoryear{Katayama et~al.}{2022}]{katayama2022domain}
\begin{botherref}
\oauthor{\bsnm{Katayama}, \binits{T.}},
\oauthor{\bsnm{Song}, \binits{T.}},
\oauthor{\bsnm{Jiang}, \binits{X.}},
\oauthor{\bsnm{Leu}, \binits{J.-S.}},
\oauthor{\bsnm{Shimamoto}, \binits{T.}}:
Domain adaptation through photorealistic enhanced images for semantic segmentation.
Mathematical Problems in Engineering
\textbf{2022}
(2022)
\end{botherref}
\endbibitem

%%% 33
\bibitem[\protect\citeauthoryear{Man and Chahl}{2022}]{man2022review}
\begin{barticle}
\bauthor{\bsnm{Man}, \binits{K.}},
\bauthor{\bsnm{Chahl}, \binits{J.}}:
\batitle{{A Review of Synthetic Image Data and Its Use in Computer Vision}}.
\bjtitle{Journal of Imaging}
\bvolume{8}(\bissue{11}),
\bfpage{310}
(\byear{2022})
\end{barticle}
\endbibitem

%%% 34
\bibitem[\protect\citeauthoryear{Yang et~al.}{2022}]{yang2022image}
\begin{botherref}
\oauthor{\bsnm{Yang}, \binits{S.}},
\oauthor{\bsnm{Xiao}, \binits{W.}},
\oauthor{\bsnm{Zhang}, \binits{M.}},
\oauthor{\bsnm{Guo}, \binits{S.}},
\oauthor{\bsnm{Zhao}, \binits{J.}},
\oauthor{\bsnm{Shen}, \binits{F.}}:
{Image Data Augmentation for Deep Learning: A Survey}.
arXiv preprint arXiv:2204.08610
(2022)
\end{botherref}
\endbibitem

%%% 35
\bibitem[\protect\citeauthoryear{Kariyappa and Qureshi}{2019}]{kariyappa2019improving}
\begin{botherref}
\oauthor{\bsnm{Kariyappa}, \binits{S.}},
\oauthor{\bsnm{Qureshi}, \binits{M.K.}}:
{Improving Adversarial Robustness of Ensembles with Diversity Training}.
arXiv preprint arXiv:1901.09981
(2019)
\end{botherref}
\endbibitem

%%% 36
\bibitem[\protect\citeauthoryear{Heusel et~al.}{2017}]{heusel2017gans}
\begin{botherref}
\oauthor{\bsnm{Heusel}, \binits{M.}},
\oauthor{\bsnm{Ramsauer}, \binits{H.}},
\oauthor{\bsnm{Unterthiner}, \binits{T.}},
\oauthor{\bsnm{Nessler}, \binits{B.}},
\oauthor{\bsnm{Hochreiter}, \binits{S.}}:
{GANs trained by a two time-scale update rule converge to a local nash equilibrium}.
Advances in neural information processing systems
\textbf{30}
(2017)
\end{botherref}
\endbibitem

%%% 37
\bibitem[\protect\citeauthoryear{Salimans et~al.}{2016}]{salimans2016improved}
\begin{botherref}
\oauthor{\bsnm{Salimans}, \binits{T.}},
\oauthor{\bsnm{Goodfellow}, \binits{I.}},
\oauthor{\bsnm{Zaremba}, \binits{W.}},
\oauthor{\bsnm{Cheung}, \binits{V.}},
\oauthor{\bsnm{Radford}, \binits{A.}},
\oauthor{\bsnm{Chen}, \binits{X.}}:
{Improved techniques for training GANs}.
Advances in neural information processing systems
\textbf{29}
(2016)
\end{botherref}
\endbibitem

%%% 38
\bibitem[\protect\citeauthoryear{Shafaei et~al.}{2016}]{ShafaeiLS16}
\begin{bchapter}
\bauthor{\bsnm{Shafaei}, \binits{A.}},
\bauthor{\bsnm{Little}, \binits{J.J.}},
\bauthor{\bsnm{Schmidt}, \binits{M.}}:
\bctitle{Play and learn: Using video games to train computer vision models}.
In: \bbtitle{Proceedings of the British Machine Vision Conference 2016, {BMVC}}
(\byear{2016}).
\bcomment{{BMVA}}
\end{bchapter}
\endbibitem

%%% 39
\bibitem[\protect\citeauthoryear{Kiefer et~al.}{2022}]{kiefer2022leveraging}
\begin{bchapter}
\bauthor{\bsnm{Kiefer}, \binits{B.}},
\bauthor{\bsnm{Ott}, \binits{D.}},
\bauthor{\bsnm{Zell}, \binits{A.}}:
\bctitle{Leveraging synthetic data in object detection on unmanned aerial vehicles}.
In: \bbtitle{2022 26th International Conference on Pattern Recognition (ICPR)},
pp. \bfpage{3564}--\blpage{3571}
(\byear{2022}).
\bcomment{IEEE}
\end{bchapter}
\endbibitem

%%% 40
\bibitem[\protect\citeauthoryear{Kerim et~al.}{2021}]{kerim2021nova}
\begin{bchapter}
\bauthor{\bsnm{Kerim}, \binits{A.}},
\bauthor{\bsnm{Aslan}, \binits{C.}},
\bauthor{\bsnm{Celikcan}, \binits{U.}},
\bauthor{\bsnm{Erdem}, \binits{E.}},
\bauthor{\bsnm{Erdem}, \binits{A.}}:
\bctitle{Nova: Rendering virtual worlds with humans for computer vision tasks}.
In: \bbtitle{Computer Graphics Forum},
vol. \bseriesno{40},
pp. \bfpage{258}--\blpage{272}
(\byear{2021}).
\bcomment{Wiley Online Library}
\end{bchapter}
\endbibitem

%%% 41
\bibitem[\protect\citeauthoryear{Lee et~al.}{2023}]{lee2023game}
\begin{barticle}
\bauthor{\bsnm{Lee}, \binits{H.}},
\bauthor{\bsnm{Jeon}, \binits{J.}},
\bauthor{\bsnm{Lee}, \binits{D.}},
\bauthor{\bsnm{Park}, \binits{C.}},
\bauthor{\bsnm{Kim}, \binits{J.}},
\bauthor{\bsnm{Lee}, \binits{D.}}:
\batitle{Game engine-driven synthetic data generation for computer vision-based safety monitoring of construction workers}.
\bjtitle{Automation in Construction}
\bvolume{155},
\bfpage{105060}
(\byear{2023})
\end{barticle}
\endbibitem

%%% 42
\bibitem[\protect\citeauthoryear{Wang et~al.}{2021}]{wang2021pixel}
\begin{barticle}
\bauthor{\bsnm{Wang}, \binits{Q.}},
\bauthor{\bsnm{Gao}, \binits{J.}},
\bauthor{\bsnm{Lin}, \binits{W.}},
\bauthor{\bsnm{Yuan}, \binits{Y.}}:
\batitle{Pixel-wise crowd understanding via synthetic data}.
\bjtitle{International Journal of Computer Vision}
\bvolume{129}(\bissue{1}),
\bfpage{225}--\blpage{245}
(\byear{2021})
\end{barticle}
\endbibitem

%%% 43
\bibitem[\protect\citeauthoryear{Rao et~al.}{2023}]{rao2023sim2realvs}
\begin{bchapter}
\bauthor{\bsnm{Rao}, \binits{Q.}},
\bauthor{\bsnm{Yu}, \binits{X.}},
\bauthor{\bsnm{Navasardyan}, \binits{S.}},
\bauthor{\bsnm{Shi}, \binits{H.}}:
\bctitle{Sim2realvs: A new benchmark for video stabilization with a strong baseline}.
In: \bbtitle{Proceedings of the IEEE/CVF Winter Conference on Applications of Computer Vision},
pp. \bfpage{5406}--\blpage{5415}
(\byear{2023})
\end{bchapter}
\endbibitem

%%% 44
\bibitem[\protect\citeauthoryear{Sun and Zheng}{2019}]{sun2019dissecting}
\begin{bchapter}
\bauthor{\bsnm{Sun}, \binits{X.}},
\bauthor{\bsnm{Zheng}, \binits{L.}}:
\bctitle{Dissecting person re-identification from the viewpoint of viewpoint}.
In: \bbtitle{Proceedings of the IEEE/CVF Conference on Computer Vision and Pattern Recognition},
pp. \bfpage{608}--\blpage{617}
(\byear{2019})
\end{bchapter}
\endbibitem

%%% 45
\bibitem[\protect\citeauthoryear{Cobbe et~al.}{2020}]{cobbe2020leveraging}
\begin{bchapter}
\bauthor{\bsnm{Cobbe}, \binits{K.}},
\bauthor{\bsnm{Hesse}, \binits{C.}},
\bauthor{\bsnm{Hilton}, \binits{J.}},
\bauthor{\bsnm{Schulman}, \binits{J.}}:
\bctitle{Leveraging procedural generation to benchmark reinforcement learning}.
In: \bbtitle{International Conference on Machine Learning},
pp. \bfpage{2048}--\blpage{2056}
(\byear{2020}).
\bcomment{PMLR}
\end{bchapter}
\endbibitem

%%% 46
\bibitem[\protect\citeauthoryear{Li et~al.}{2024}]{li2024bridging}
\begin{bchapter}
\bauthor{\bsnm{Li}, \binits{A.}},
\bauthor{\bsnm{Wu}, \binits{J.}},
\bauthor{\bsnm{Liu}, \binits{Y.}},
\bauthor{\bsnm{Li}, \binits{L.}}:
\bctitle{Bridging the synthetic-to-authentic gap: Distortion-guided unsupervised domain adaptation for blind image quality assessment}.
In: \bbtitle{Proceedings of the IEEE/CVF Conference on Computer Vision and Pattern Recognition},
pp. \bfpage{28422}--\blpage{28431}
(\byear{2024})
\end{bchapter}
\endbibitem

%%% 47
\bibitem[\protect\citeauthoryear{Hao et~al.}{2024}]{hao2024synthetic}
\begin{botherref}
\oauthor{\bsnm{Hao}, \binits{S.}},
\oauthor{\bsnm{Han}, \binits{W.}},
\oauthor{\bsnm{Jiang}, \binits{T.}},
\oauthor{\bsnm{Li}, \binits{Y.}},
\oauthor{\bsnm{Wu}, \binits{H.}},
\oauthor{\bsnm{Zhong}, \binits{C.}},
\oauthor{\bsnm{Zhou}, \binits{Z.}},
\oauthor{\bsnm{Tang}, \binits{H.}}:
Synthetic data in ai: Challenges, applications, and ethical implications.
arXiv preprint arXiv:2401.01629
(2024)
\end{botherref}
\endbibitem

%%% 48
\bibitem[\protect\citeauthoryear{Ali et~al.}{2023}]{ali2023leveraging}
\begin{bchapter}
\bauthor{\bsnm{Ali}, \binits{H.}},
\bauthor{\bsnm{Gr{\"o}nlund}, \binits{C.}},
\bauthor{\bsnm{Shah}, \binits{Z.}}:
\bctitle{{Leveraging GANs for Data Scarcity of COVID-19: Beyond the Hype}}.
In: \bbtitle{Proceedings of the IEEE/CVF Conference on Computer Vision and Pattern Recognition},
pp. \bfpage{659}--\blpage{667}
(\byear{2023})
\end{bchapter}
\endbibitem

%%% 49
\bibitem[\protect\citeauthoryear{Torfi et~al.}{2022}]{torfi2022differentially}
\begin{barticle}
\bauthor{\bsnm{Torfi}, \binits{A.}},
\bauthor{\bsnm{Fox}, \binits{E.A.}},
\bauthor{\bsnm{Reddy}, \binits{C.K.}}:
\batitle{{Differentially private synthetic medical data generation using convolutional GANs}}.
\bjtitle{Information Sciences}
\bvolume{586},
\bfpage{485}--\blpage{500}
(\byear{2022})
\end{barticle}
\endbibitem

%%% 50
\bibitem[\protect\citeauthoryear{Al~Khalil et~al.}{2023}]{al2023usability}
\begin{barticle}
\bauthor{\bsnm{Al~Khalil}, \binits{Y.}},
\bauthor{\bsnm{Amirrajab}, \binits{S.}},
\bauthor{\bsnm{Lorenz}, \binits{C.}},
\bauthor{\bsnm{Weese}, \binits{J.}},
\bauthor{\bsnm{Pluim}, \binits{J.}},
\bauthor{\bsnm{Breeuwer}, \binits{M.}}:
\batitle{{On the Usability of Synthetic Data for Improving the Robustness of Deep Learning-based Segmentation of Cardiac Magnetic Resonance Images}}.
\bjtitle{Medical Image Analysis}
\bvolume{84},
\bfpage{102688}
(\byear{2023})
\end{barticle}
\endbibitem

%%% 51
\bibitem[\protect\citeauthoryear{Li et~al.}{2023}]{li2023lift3d}
\begin{bchapter}
\bauthor{\bsnm{Li}, \binits{L.}},
\bauthor{\bsnm{Lian}, \binits{Q.}},
\bauthor{\bsnm{Wang}, \binits{L.}},
\bauthor{\bsnm{Ma}, \binits{N.}},
\bauthor{\bsnm{Chen}, \binits{Y.-C.}}:
\bctitle{{Lift3D: Synthesize 3D Training Data by Lifting 2D GAN to 3D Generative Radiance Field}}.
In: \bbtitle{Proceedings of the IEEE/CVF Conference on Computer Vision and Pattern Recognition},
pp. \bfpage{332}--\blpage{341}
(\byear{2023})
\end{bchapter}
\endbibitem

%%% 52
\bibitem[\protect\citeauthoryear{Kodali et~al.}{2017}]{kodali2017convergence}
\begin{botherref}
\oauthor{\bsnm{Kodali}, \binits{N.}},
\oauthor{\bsnm{Abernethy}, \binits{J.}},
\oauthor{\bsnm{Hays}, \binits{J.}},
\oauthor{\bsnm{Kira}, \binits{Z.}}:
{On convergence and stability of GANs}.
arXiv preprint arXiv:1705.07215
(2017)
\end{botherref}
\endbibitem

%%% 53
\bibitem[\protect\citeauthoryear{Gulrajani et~al.}{2017}]{gulrajani2017improved}
\begin{botherref}
\oauthor{\bsnm{Gulrajani}, \binits{I.}},
\oauthor{\bsnm{Ahmed}, \binits{F.}},
\oauthor{\bsnm{Arjovsky}, \binits{M.}},
\oauthor{\bsnm{Dumoulin}, \binits{V.}},
\oauthor{\bsnm{Courville}, \binits{A.C.}}:
{Improved training of Wasserstein GANs}.
Advances in neural information processing systems
\textbf{30}
(2017)
\end{botherref}
\endbibitem

%%% 54
\bibitem[\protect\citeauthoryear{Yang et~al.}{2023}]{yang2023diffusion}
\begin{barticle}
\bauthor{\bsnm{Yang}, \binits{L.}},
\bauthor{\bsnm{Zhang}, \binits{Z.}},
\bauthor{\bsnm{Song}, \binits{Y.}},
\bauthor{\bsnm{Hong}, \binits{S.}},
\bauthor{\bsnm{Xu}, \binits{R.}},
\bauthor{\bsnm{Zhao}, \binits{Y.}},
\bauthor{\bsnm{Zhang}, \binits{W.}},
\bauthor{\bsnm{Cui}, \binits{B.}},
\bauthor{\bsnm{Yang}, \binits{M.-H.}}:
\batitle{Diffusion models: A comprehensive survey of methods and applications}.
\bjtitle{ACM Computing Surveys}
\bvolume{56}(\bissue{4}),
\bfpage{1}--\blpage{39}
(\byear{2023})
\end{barticle}
\endbibitem

%%% 55
\bibitem[\protect\citeauthoryear{Shipard et~al.}{2023}]{shipard2023diversity}
\begin{bchapter}
\bauthor{\bsnm{Shipard}, \binits{J.}},
\bauthor{\bsnm{Wiliem}, \binits{A.}},
\bauthor{\bsnm{Thanh}, \binits{K.N.}},
\bauthor{\bsnm{Xiang}, \binits{W.}},
\bauthor{\bsnm{Fookes}, \binits{C.}}:
\bctitle{Diversity is definitely needed: Improving model-agnostic zero-shot classification via stable diffusion}.
In: \bbtitle{Proceedings of the IEEE/CVF Conference on Computer Vision and Pattern Recognition},
pp. \bfpage{769}--\blpage{778}
(\byear{2023})
\end{bchapter}
\endbibitem

%%% 56
\bibitem[\protect\citeauthoryear{Akrout et~al.}{2023}]{akrout2023diffusion}
\begin{botherref}
\oauthor{\bsnm{Akrout}, \binits{M.}},
\oauthor{\bsnm{Gyepesi}, \binits{B.}},
\oauthor{\bsnm{Holl{\'o}}, \binits{P.}},
\oauthor{\bsnm{Po{\'o}r}, \binits{A.}},
\oauthor{\bsnm{Kincs{\H{o}}}, \binits{B.}},
\oauthor{\bsnm{Solis}, \binits{S.}},
\oauthor{\bsnm{Cirone}, \binits{K.}},
\oauthor{\bsnm{Kawahara}, \binits{J.}},
\oauthor{\bsnm{Slade}, \binits{D.}},
\oauthor{\bsnm{Abid}, \binits{L.}}, et al.:
{Diffusion-based Data Augmentation for Skin Disease Classification: Impact Across Original Medical Datasets to Fully Synthetic Images}.
arXiv preprint arXiv:2301.04802
(2023)
\end{botherref}
\endbibitem

%%% 57
\bibitem[\protect\citeauthoryear{Wang et~al.}{2004}]{wang2004image}
\begin{barticle}
\bauthor{\bsnm{Wang}, \binits{Z.}},
\bauthor{\bsnm{Bovik}, \binits{A.C.}},
\bauthor{\bsnm{Sheikh}, \binits{H.R.}},
\bauthor{\bsnm{Simoncelli}, \binits{E.P.}}:
\batitle{Image quality assessment: from error visibility to structural similarity}.
\bjtitle{IEEE transactions on image processing}
\bvolume{13}(\bissue{4}),
\bfpage{600}--\blpage{612}
(\byear{2004})
\end{barticle}
\endbibitem

%%% 58
\bibitem[\protect\citeauthoryear{Santangelo et~al.}{2024}]{santangelo2024synthcheck}
\begin{bchapter}
\bauthor{\bsnm{Santangelo}, \binits{G.}},
\bauthor{\bsnm{Nicora}, \binits{G.}},
\bauthor{\bsnm{Bellazzi}, \binits{R.}},
\bauthor{\bsnm{Dagliati}, \binits{A.}}, \betal:
\bctitle{Synthcheck: A dashboard for synthetic data quality assessment.}
In: \bbtitle{BIOSTEC (2)},
pp. \bfpage{246}--\blpage{256}
(\byear{2024})
\end{bchapter}
\endbibitem

%%% 59
\bibitem[\protect\citeauthoryear{Breuer et~al.}{2024}]{breuer2024validating}
\begin{barticle}
\bauthor{\bsnm{Breuer}, \binits{T.}},
\bauthor{\bsnm{Fuhr}, \binits{N.}},
\bauthor{\bsnm{Schaer}, \binits{P.}}:
\batitle{Validating synthetic usage data in living lab environments}.
\bjtitle{ACM Journal of Data and Information Quality}
\bvolume{16}(\bissue{1}),
\bfpage{1}--\blpage{33}
(\byear{2024})
\end{barticle}
\endbibitem

%%% 60
\bibitem[\protect\citeauthoryear{Vallevik et~al.}{2024}]{vallevik2024can}
\begin{botherref}
\oauthor{\bsnm{Vallevik}, \binits{V.B.}},
\oauthor{\bsnm{Babic}, \binits{A.}},
\oauthor{\bsnm{Marshall}, \binits{S.E.}},
\oauthor{\bsnm{Severin}, \binits{E.}},
\oauthor{\bsnm{Br{\o}gger}, \binits{H.M.}},
\oauthor{\bsnm{Alagaratnam}, \binits{S.}},
\oauthor{\bsnm{Edwin}, \binits{B.}},
\oauthor{\bsnm{Veeraragavan}, \binits{N.R.}},
\oauthor{\bsnm{Befring}, \binits{A.K.}},
\oauthor{\bsnm{Nyg{\aa}rd}, \binits{J.F.}}:
{Can I trust my fake data--A comprehensive quality assessment framework for synthetic tabular data in healthcare}.
International Journal of Medical Informatics,
105413
(2024)
\end{botherref}
\endbibitem

%%% 61
\bibitem[\protect\citeauthoryear{Lucic et~al.}{2018}]{lucic2018gans}
\begin{botherref}
\oauthor{\bsnm{Lucic}, \binits{M.}},
\oauthor{\bsnm{Kurach}, \binits{K.}},
\oauthor{\bsnm{Michalski}, \binits{M.}},
\oauthor{\bsnm{Gelly}, \binits{S.}},
\oauthor{\bsnm{Bousquet}, \binits{O.}}:
{Are GANs created equal? A large-scale study}.
Advances in neural information processing systems
\textbf{31}
(2018)
\end{botherref}
\endbibitem

%%% 62
\bibitem[\protect\citeauthoryear{Borji}{2019}]{borji2019pros}
\begin{barticle}
\bauthor{\bsnm{Borji}, \binits{A.}}:
\batitle{{Pros and cons of GAN evaluation measures}}.
\bjtitle{Computer vision and image understanding}
\bvolume{179},
\bfpage{41}--\blpage{65}
(\byear{2019})
\end{barticle}
\endbibitem

%%% 63
\bibitem[\protect\citeauthoryear{Szegedy et~al.}{2016}]{szegedy2016rethinking}
\begin{bchapter}
\bauthor{\bsnm{Szegedy}, \binits{C.}},
\bauthor{\bsnm{Vanhoucke}, \binits{V.}},
\bauthor{\bsnm{Ioffe}, \binits{S.}},
\bauthor{\bsnm{Shlens}, \binits{J.}},
\bauthor{\bsnm{Wojna}, \binits{Z.}}:
\bctitle{Rethinking the inception architecture for computer vision}.
In: \bbtitle{Proceedings of the IEEE Conference on Computer Vision and Pattern Recognition},
pp. \bfpage{2818}--\blpage{2826}
(\byear{2016})
\end{bchapter}
\endbibitem

%%% 64
\bibitem[\protect\citeauthoryear{Barratt and Sharma}{2018}]{barratt2018note}
\begin{botherref}
\oauthor{\bsnm{Barratt}, \binits{S.}},
\oauthor{\bsnm{Sharma}, \binits{R.}}:
A note on the inception score.
arXiv preprint arXiv:1801.01973
(2018)
\end{botherref}
\endbibitem

%%% 65
\bibitem[\protect\citeauthoryear{Ravuri and Vinyals}{2019}]{ravuri2019seeing}
\begin{botherref}
\oauthor{\bsnm{Ravuri}, \binits{S.}},
\oauthor{\bsnm{Vinyals}, \binits{O.}}:
{Seeing is not necessarily believing: Limitations of BigGANs for data augmentation}
(2019)
\end{botherref}
\endbibitem

%%% 66
\bibitem[\protect\citeauthoryear{Pambrun and Noumeir}{2015}]{pambrun2015limitations}
\begin{bchapter}
\bauthor{\bsnm{Pambrun}, \binits{J.-F.}},
\bauthor{\bsnm{Noumeir}, \binits{R.}}:
\bctitle{Limitations of the ssim quality metric in the context of diagnostic imaging}.
In: \bbtitle{2015 IEEE International Conference on Image Processing (ICIP)},
pp. \bfpage{2960}--\blpage{2963}
(\byear{2015}).
\bcomment{IEEE}
\end{bchapter}
\endbibitem

%%% 67
\bibitem[\protect\citeauthoryear{Huynh-Thu and Ghanbari}{2008}]{huynh2008scope}
\begin{barticle}
\bauthor{\bsnm{Huynh-Thu}, \binits{Q.}},
\bauthor{\bsnm{Ghanbari}, \binits{M.}}:
\batitle{{Scope of validity of PSNR in image/video quality assessment}}.
\bjtitle{Electronics letters}
\bvolume{44}(\bissue{13}),
\bfpage{800}--\blpage{801}
(\byear{2008})
\end{barticle}
\endbibitem

%%% 68
\bibitem[\protect\citeauthoryear{Scholz et~al.}{2023}]{scholz2023metrics}
\begin{botherref}
\oauthor{\bsnm{Scholz}, \binits{D.}},
\oauthor{\bsnm{Wiestler}, \binits{B.}},
\oauthor{\bsnm{Rueckert}, \binits{D.}},
\oauthor{\bsnm{Menten}, \binits{M.J.}}:
{Metrics to Quantify Global Consistency in Synthetic Medical Images}.
arXiv preprint arXiv:2308.00402
(2023)
\end{botherref}
\endbibitem

%%% 69
\bibitem[\protect\citeauthoryear{Mahon and Lukasiewicz}{2022}]{mahon2022measuring}
\begin{botherref}
\oauthor{\bsnm{Mahon}, \binits{L.}},
\oauthor{\bsnm{Lukasiewicz}, \binits{T.}}:
{Measuring Image Complexity as a Discrete Hierarchy using MDL Clustering}
(2022)
\end{botherref}
\endbibitem

%%% 70
\bibitem[\protect\citeauthoryear{Kullback and Leibler}{1951}]{kullback1951information}
\begin{barticle}
\bauthor{\bsnm{Kullback}, \binits{S.}},
\bauthor{\bsnm{Leibler}, \binits{R.A.}}:
\batitle{On information and sufficiency}.
\bjtitle{The annals of mathematical statistics}
\bvolume{22}(\bissue{1}),
\bfpage{79}--\blpage{86}
(\byear{1951})
\end{barticle}
\endbibitem

%%% 71
\bibitem[\protect\citeauthoryear{Rombach et~al.}{2022}]{Rombach_2022_CVPR}
\begin{bchapter}
\bauthor{\bsnm{Rombach}, \binits{R.}},
\bauthor{\bsnm{Blattmann}, \binits{A.}},
\bauthor{\bsnm{Lorenz}, \binits{D.}},
\bauthor{\bsnm{Esser}, \binits{P.}},
\bauthor{\bsnm{Ommer}, \binits{B.}}:
\bctitle{{High-Resolution Image Synthesis With Latent Diffusion Models}}.
In: \bbtitle{Proceedings of the IEEE/CVF Conference on Computer Vision and Pattern Recognition (CVPR)},
pp. \bfpage{10684}--\blpage{10695}
(\byear{2022})
\end{bchapter}
\endbibitem

\end{thebibliography}
% %% if required, the content of .bbl file can be included here once bbl is generated
% %%\input sn-article.bbl

\end{document}